\documentclass[a4paper,twoside]{article}

\usepackage{graphicx}
\usepackage{makeidx}

\usepackage{color}
\usepackage{epsfig}
\usepackage{subfigure}
\usepackage{calc}
\usepackage{amssymb}
\usepackage{amstext}
\usepackage{amsmath}
\usepackage{amsthm}
\usepackage{multicol}
\usepackage{pslatex}
\usepackage{apalike}
\usepackage{SCITEPRESS}
\usepackage[small]{caption}

\newcommand{\imgsize}{0.24}
\newcommand{\imgsizezoom}{0.24}
\newcommand{\xlev}{130}
\newcommand{\xlevtitle}{188}
\newcommand{\imWidth}{.30}
\newcommand{\brar}[1]{\left(#1\right)}
\newcommand{\brae}[1]{\left[#1\right]}

\newcommand{\bras}[1]{\left\langle #1\right\rangle}

\newcommand{\vc}{\textit{\textbf{c}}}
\newcommand{\vB}{\textit{\textbf{B}}}
\newcommand{\vS}{\textit{\textbf{S}}}
\newcommand{\vn}{\textit{\textbf{n}}}
\newcommand{\eg}{e.g.\ }
\newcommand{\ie}{i.e.\ }
\newcommand{\dive}{$\mbox{div}$}

\begin{document}

\title{Using Channel Representations in Regularization Terms: A Case Study on Image Diffusion}

\author{\authorname{Christian Heinemann\sup{1}, Freddie {\AA}str\"om\sup{2}, George Baravdish\sup{3}, Kai Krajsek\sup{1}, Michael Felsberg\sup{2} \\ and Hanno Scharr\sup{1}}
\affiliation{\sup{1}IBG-2: Plant Sciences, Forschungszentrum J\"ulich, 52425 J\"ulich, Germany}
\affiliation{\sup{2}Department of Electrical Engineering, Link\"oping University, SE-581 83 Link\"oping, Sweden}
\affiliation{\sup{3}Department of Science and Technology, Link\"oping University, SE-601 74 Norrk\"oping, Sweden}
\email{\{c.heinemann, k.krajsek, h.scharr\}@fz-juelich.de, \{freddie.astrom, michael.felsberg, george.baravdish\}@liu.se}
}


\abstract{
In this work we propose a novel non-linear diffusion filtering approach for images based on their channel representation. To derive
the diffusion update scheme we formulate a novel energy functional using
a soft-histogram representation of image pixel neighborhoods obtained
from the channel encoding. The resulting Euler-Lagrange equation yields
a non-linear robust diffusion scheme with additional weighting terms
 stemming from the channel representation which steer the diffusion process. We apply this novel energy formulation to image reconstruction
 problems, showing good performance in the presence of mixtures of Gaussian and impulse-like noise, \eg missing data. In denoising experiments of
 common scalar-valued images our approach performs competitive compared to other diffusion schemes as well as state-of-the-art denoising
 methods for the considered noise types.}

\onecolumn \maketitle \normalsize \vfill

\section{Introduction}

The channel representation \cite{gg2000g} is a special, lossless soft-histogram, in contrast to other histograms. Encoding a single value in a channel representation, the value can be accurately reconstructed. Using the channel representation allows for simple outlier-removing smoothing, so called channel smoothing \cite{ffs05}, by essentially for each pixel encoding the values of its local neighborhood into a soft-histogram and assigning the decoded value of the dominant mode to it, \ie the sub-'bin'-position of the maximum value in the soft-histogram.

This paper studies the nature and performance of regularization terms penalizing the gradient of a channel-smoothed signal. The rational behind this is the observation, that in a robust diffusion scheme \cite{Perona90scale-spaceand}, which can be derived from a regularization term penalizing the gradient of a signal, the smoothing process is stopped or reduced at edges, \ie high gradient values. An outlier may erroneously be detected as edge and by this not smoothed away. Not considering the outlier in the penalizing term thus is expected to result in still high smoothing strengths at outlier positions, reducing their visibility.

The natural test case for such a regularization is its use without a data term, \ie image diffusion not regularization. Further, as the robustness to outliers is of interest, one natural application is image reconstruction in the presence of mixed noises. We use mixtures of Gaussian and impulse-like noise.

We do not expect to derive the novel best denoising scheme for gray value image reconstruction and our experiments show that state-of-the-art denoising schemes do a better job here. However, this additional robustification by channel smoothing can easily be transferred to application domains, where diffusion schemes are used in practice. Diffusion schemes have been derived \eg for color \cite{KimmelMS98},
vector \cite{TschumperleD05}, or matrix-valued data \cite{Burgeth07}. They are applied for regularization of DTMRI \cite{Kraj} or high angular resolution diffusion imaging (HARDI) data \cite{Goh20111181}. Instead of a scalar-valued diffusivities, tensorbased diffusion schemes use tensor-valued diffusivities describing single or multiple orientations in the image to drive the diffusion process \cite{Weickert96anisotropicdiffusion,ScharrDAGM2006}.

Crucial for the performance of non-linear diffusion is an adaptation of the diffusivity, \ie the local smoothing strength, to image structure. Typically image structure is measured as the Euclidean norm of the local image gradient. It is transformed into a diffusivity by means of a \textit{edge-stopping function}, assigning small diffusivities to locations with high gradient and vice versa. The exact choice of the edge-stopping function has been shown to be equivalent to the choice of error norm in robust statistics \cite{Black1998} and can be learned from image statistics \cite{ZhuMumford,RothBlack}. %Add perhaps sff03 here.

If outliers are present in the data, \eg as salt-and-pepper noise, or dropouts, other reconstruction methods than diffusion are usually applied, able to remove outliers completely, \eg median filtering (see \eg \cite{GonzalezWoods}), or channel smoothing \cite{ffs05} selecting the maximum mode of the local value distribution. Channel smoothing (CS) averages not only by applying a spatial window, but also windows in the value domain. However, the value domain window is not centered at the value of the currently processed pixel, in contrast to bilateral filtering \cite{Tomasi98bilateralfiltering}, but centered at the maximum of the local value distribution. By this, it removes clear outliers and interpolates from neighbors. If Gaussian noise is present in the data, also a part of the inliers are lost due to the value domain window and not considered in the value reconstruction. Consequently the reconstructed gray value is less efficiently denoised as if all inliers were considered.

\begin{figure*}[tb]
  \includegraphics[width=1\linewidth]{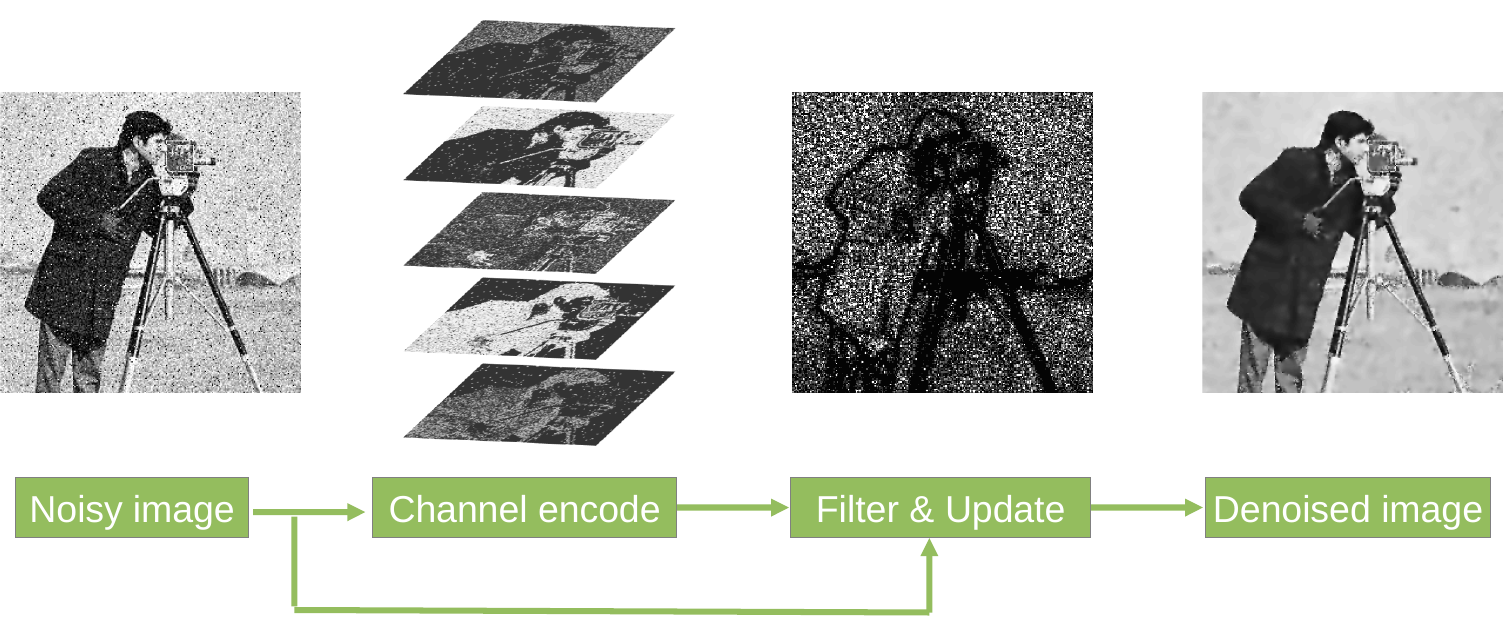}
  \caption{Overview of proposed method. Contrast is increased in "Filter \& Update" for a better visualization.}
  \label{fig:overview}
\end{figure*}
\subsection{Main contributions} 
The core idea of this paper is to drive a non-linear diffusion process using the CS result as image structure information. CS alone gives poor denoising results for medium to high Gaussian noise levels. In this case, non-linear diffusion can yield higher signal-to-noise-ratio (SNR) as it can average over all available data. However, usual non-linear diffusion which is driven by the local gradient stops at clear outliers, not suppressing them. CS removes outliers while preserving edge location and roughly edge strength. Thus driving a non-linear diffusion by CS allows to oversmooth -- not remove -- outliers. We therefore expect to gain SNR compared to plain CS when a considerable amount of Gaussian noise is present and still suppress outliers such that they are less visible in the image.
To derive the CS-based diffusion term, we define an energy term penalizing the gradient of a channel encoded image.

The schematic diffusion approach is illustrated in Fig.~\ref{fig:overview}. The noisy image is decomposed into its corresponding channel representation. Within the channel space we determine a local neighborhood which best represents the data sample in the image space, thus we guide a diffusion process such that Gaussian and impulse noise is reduced which is indicated by the white spots in the "Filter \& Update" step. Black indicates no update, so important structure and edges in the image are preserved.

Clearly, our diffusion process cannot {\em remove} outliers from data, but strongly smooth impulse-like outliers. It is therefore well adapted to improve the visual impression and preserving the structure of reconstructed images, measured by SSIM \cite{1284395} in our experiments. Due to the mean-preserving property of diffusion filtering, bias-free improvement of the reconstructed data values can only be hoped for, when zero-mean noise is present.

\section{Background theory }

In order to derive the novel non-linear channel-based diffusion (NLCD) scheme in the subsequent section we first recall image diffusion derived from an energy minimization, before the framework of channel representation and channel smoothing is briefly described. 

\subsection{Regularization and Image diffusion}
\label{linearchanneldiffusion}
 Let $u(x,y) : \Omega \subset \mathbb{R}^2\rightarrow\mathbb{R}$ be a scalar valued image defined on a subset of $\mathbb{R}^2$. The variational approach to image regularization or image diffusion\footnote{%
Image regularization and image diffusion are closely related. Image regularization is a boundary value problem, equivalent to heat diffusion with source terms in physics. Image diffusion is equivalent to heat diffusion without sources, formulated as initial value problem, starting with the measured, noisy image. In both cases, the regularization term defines the smoothing process leading to noise suppression. For simplicity, we call this process {\em diffusion} in both cases, as usual in physics. For details of the very close relation between regularization and diffusion see \cite{Scherzer98relationsbetween}.
} is to consider an energy functional of the form 
\begin{equation}
E(u) = \frac{1}{2}\int_\Omega (u-u^0)^2 \; dxdy + \lambda R(u) \enspace ,
\end{equation}
where $R(u)$ is a regularization term, $\lambda \in \mathbb{R}, \lambda > 0$ controls the influence of the regularization and $u^0$ is the observed image. In particular let $R(u)$ be selected as
\begin{equation}
  R(u) =  \frac{1}{2} \int_\Omega |\nabla u|^2 \; dxdy \enspace ,
  \label{eq:linear_diff_R}
\end{equation}
where the gradient operator is defined as $\nabla = \begin{pmatrix} \partial_x, & \partial_y \end{pmatrix}^t$ and $\left|\cdot\right|$ denotes the Euclidean norm. Then the functional $E(u)$ defines linear image regularization:
To minimize $E(u)$ one can compute the variational derivative and by assuming Neumann boundary conditions $\bras{\nabla u, \vn}=0$ for a boundary element $\vn$ of $\partial \Omega$, we obtain the Euler-Lagrange (EL) equation described by the PDE
\begin{equation}
\left\{\begin{array}{rcl}
u-u^0 - \lambda \Delta u &=& 0 \\
\bras{\nabla u, \vn} & = & 0  .
\end{array}\right. 
\label{updscheme_linear} 
\end{equation}
The corresponding linear diffusion without data term reads
\begin{equation}
\left\{\begin{array}{rcl}
\partial_t u &=& \Delta u \\
\bras{\nabla u, \vn} & = & 0  .
\end{array}\right. 
\label{linear_diffusion} 
\end{equation}

\subsection{Channel representation}
\label{channelrep}
\label{sec:channels}
\begin{figure}
\centering
  \includegraphics[width=0.8\linewidth]{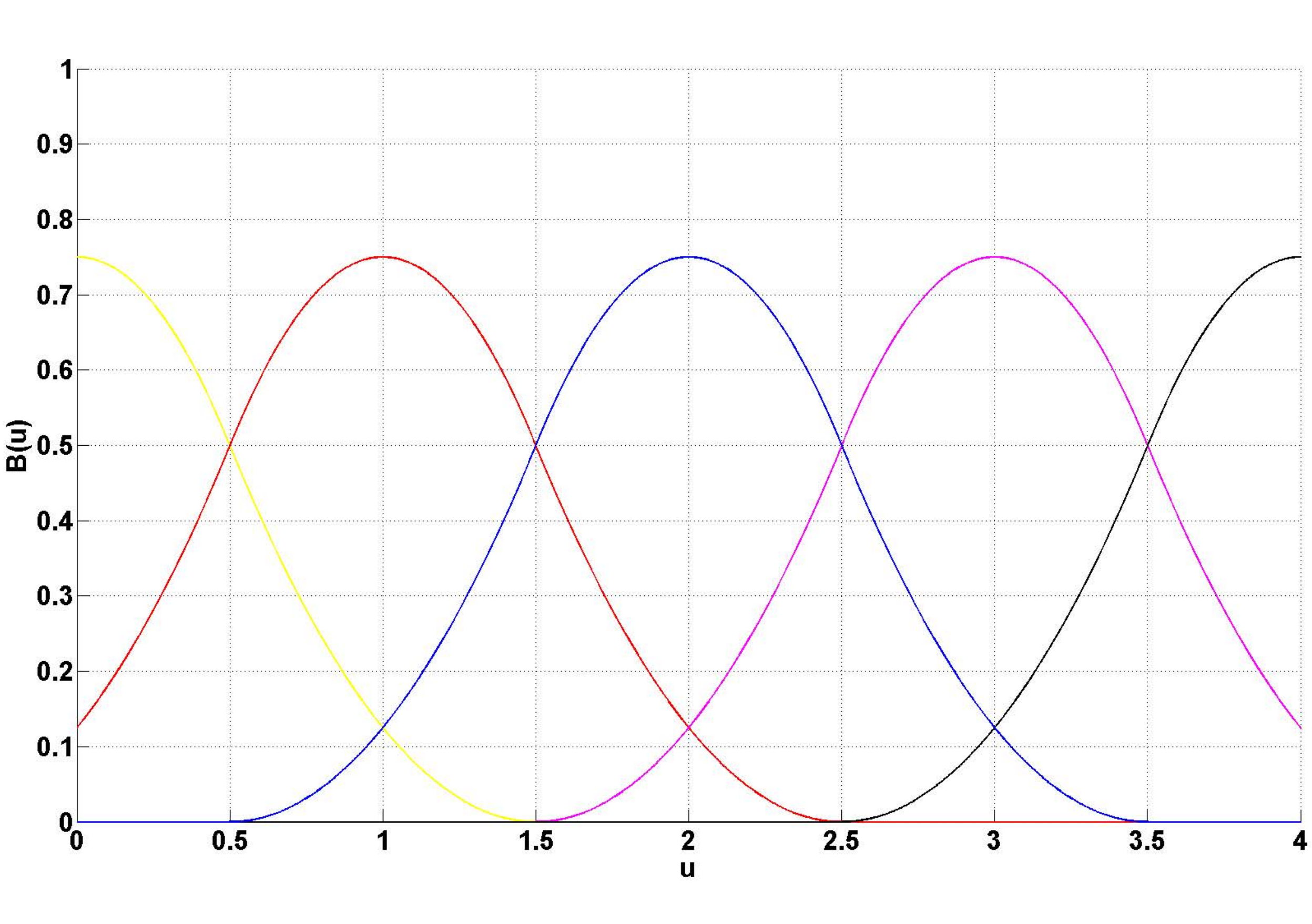}
  \caption{Example for equidistantly distributed B-Spline basis functions. Five channels are used here.}
  \label{fig:basisfunctions}
\end{figure}
This section introduces channel representations as an approximative density estimator as described in~\cite{ffs05}. Channel representations are basically soft-histograms, i.e., histograms where samples are not exclusively pooled to the closest bin center, but to several bins with a weight depending on the distance to the respective bin center. The 'bins' are called 'channels'. The binning operator or density mapping function is called a \textit{basis function}, $B(u)$, and is usually non-negative (as densities are non-negative), has compact support and is smooth (for stability reasons). The measure induced by the mapping to channel representations should be position independent in order to avoid an unwanted bias. In this paper, quadratic B-splines are used as basis functions:
\begin{align}
  B(u) & = 
  \left\{\begin{array}{lcl}
      \dfrac{3}{4} - u^2  & & 0 \leq |u| \leq \dfrac{1}{2} \\
      \dfrac{1}{2}\left (\dfrac{3}{4} - |u| \right )^2  & & \dfrac{1}{2} < |u| \leq \dfrac{3}{2} \\
      0  & & \mbox{Otherwise} .
    \end{array}\right. 
\end{align}
Without loss of generality we assume that $u(x)\in [0, N-1]$ before the signal can be encoded into $N$ channels. Otherwise we linearly transform $u$ to that interval. Figure \ref{fig:basisfunctions} depicts the basis functions in the case of 5 channels. Let $\vc \in \mathbb{R}^N$, $N>0$, be an equidistantly distributed grid over the image range of $u$. We call $\vc = (c_1, \ldots , c_N)^t$ \textit{channel center vector}. Then we obtain the quadratic B-Spline channel representation  $B_i = B(u(x,y) - c_i)$ 
where $\vB = (B_1, \ldots, B_N), i=1, \ldots,N$ is called \textit{channel vector}.

In channel smoothing \cite{ffs05} the channel vectors are spatially averaged using a Gaussian kernel $w$ resulting in 
\begin{equation}
\tilde{B}_i(x, y) = w \ast B_i(x, y) .
\label{eq.chanSmooth}
\end{equation}
Reconstructing a value $u$ from the channel representation can be done using a linear combination of \textbf{all} channel vector components yielding a linear decoding. 
To obtain a \textit{robust} decoding scheme only a \textbf{subset} of the channel components is used \cite{ffs05}. In this work a window of size $3$ around the maximum mode location of the channel vector is computed \ie  
\begin{equation}
  l=\operatorname*{argmax}_k \sum\hspace{-2mm}{\phantom{\big)}}_{i=k-1}^{k+1} \tilde{B}_i(x,y)\enspace 
	\label{argmax}
\end{equation}
with $k=2,\ldots , N-1$. Then the \emph{robust} decoding scheme, also used in the framework of channel smoothing \cite{ffs05}, reads
\begin{equation}
  \hat{u}(x,y) = \sum_{i=l-1}^{l+1}  c_i\tilde{B}_i(x,y)   \enspace . 
	\label{chansmoothedu}
\end{equation} 
where we assume that the coeffcients $\tilde{B}_i(x,y)$ sum up to one, if not, we normalize them to do so. Note that the choice of $l$ in (\ref{argmax}) depends on continuously differentiable basis functions such that the local maxima are continuous functions of the input values. The order of local maxima depends on the clusters of the inputs, but this is desirable since this reflects \eg the non-stationarity across edges.

\section{Image diffusion with the channel framework}

In this section we introduce an energy functional combining the framework of diffusion filtering and channel representation in order to first derive linear channel-based diffusion (LCD) before extending it to the non-linear case (NLCD).

\subsection{Introducing channel representation to linear image diffusion }
\begin{figure}
  \centering
	\begin{picture}(300,110)
  \put(0,-5)     {\includegraphics[width=0.47\linewidth]{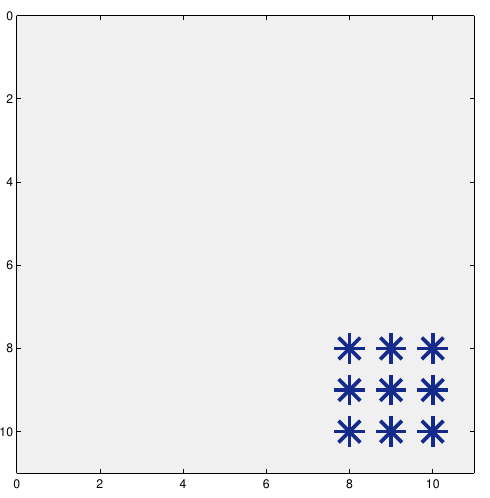}}
	\put(109,1)   {\includegraphics[width=0.45\linewidth]{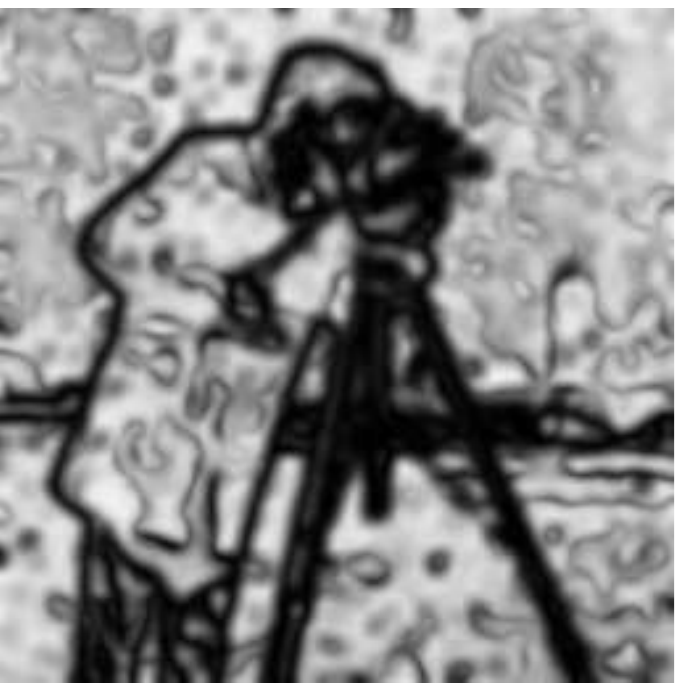}}
	\put(5,102)   {a)}
	\put(99,102) {b)}
	\end{picture}

	\centering
  \caption{a) Typical structure of $\vS(u)$ for a certain spatial position. The size of the matrix is equal to the number of channels. In this example 11 channels are used and non-zero entries are centered around channel 9. b) Structure of $\vc^t\vS(u) \vc$ for the cameraman image and a Gaussian kernel $w$ with standard deviation 3. Black indicates low and white high values close to 1. }
\label{fig:StructureS}
\end{figure}

In order to enable diffusion methods to oversmooth impulse noise, we define the regularization term $R(u)$ using the \emph{channel representation} as described in the previous section. 
We encode $u$ into the channel space using (\ref{chansmoothedu}) thus leading to the regularization term
\begin{equation}
  R(u) = \int_\Omega |\nabla \sum_{i =l-1}^{l+1} c_i \tilde{B}_i(x,y) |^2 \ dxdy \enspace ,
  \label{nonlinenergy} 
\end{equation}
where $l$ is defined in (\ref{argmax}) and $\tilde{B}$ are the smoothed channel weights as in (\ref{eq.chanSmooth}). With notational abuse, to simplify notation, we define in the subsequent derivation
\begin{equation}
  \sum_{i =l-1}^{l+1} c_i \tilde{B}_i(x,y) = \vc^t \tilde{\vB} \enspace ,
\end{equation}
where the 3-box windowing is implicitly included in $\tilde{\vB}(u)$. The integrand of $R(u)$ can then be written in vector-matrix notation as $|\nabla(\vc^t \tilde{\vB})|^2 $ which yields
\begin{equation}
  R(u) =  \int_\Omega \vc^t \brae{(\partial_x \tilde{\vB}) (\partial_x \tilde{\vB})^t +  (\partial_y \tilde{\vB}) (\partial_y \tilde{\vB})^t} \vc \ dxdy \enspace .
  \label{Rofu}
\end{equation}
To find the EL-equation of \eqref{Rofu} we compute the variational derivative, denoted as $ R_uv = \left.\frac{\partial R(u+\epsilon v)}{\partial \epsilon}\right|_{\epsilon = 0}$, in the direction of $v$ with respect to $u$ where $v\in C^2(\Omega)$ is a non-zero testfunction.
We split the integral (\ref{Rofu}) into two integrals calculating the $x$- and $y$-component separately and get
\begin{align*}
  R_u v & = \frac{\partial}{\partial \epsilon} \left[ \int_\Omega \vc^t  \tilde{\vB}'(u+\varepsilon v)\tilde{\vB}'(u+\varepsilon v)^t \right.\\
	& \hspace{17mm} \cdot\left.\left.\brar{\partial_x u + \epsilon \partial_x v}^2 \vc \ dxdy \ \right]\right|_{\epsilon = 0} \\
  & =  \int_\Omega  \vc^t \Big ( \frac{\partial}{\partial \epsilon}\left.\brae{ \tilde{\vB}'(u+\varepsilon v)\tilde{\vB}'(u+\varepsilon v)^t}\right|_{\epsilon = 0} \Big ) \vc \brar{\partial_x u}^2 \\
  & \hspace{15mm} + 2 \vc^t \Big ( \tilde{\vB}'(u) \tilde{\vB}'(u)^t\partial_x u \partial_x v \Big )\vc  \ dxdy \\
	& = \int_\Omega \vc^t\brar{\tilde{\vB}''(u)\tilde{\vB}'(u)^t + \tilde{\vB}'(u)\tilde{\vB}''(u)^t}\vc\brar{\partial_x u}^2 v \ dxdy  \\
	& \hspace{15mm} - 2 \int_\Omega \partial_x \brar{\vc^t \tilde{\vB}'(u) \tilde{\vB}'(u)^t \vc \partial_x u}  v \ dxdy \enspace .
  \label{eq:2}
\end{align*}
The same holds for the $y$-component. For the last equation we applied Green's formula and Neumann boundary conditions to get rid of the $\partial_x v$ term which cannot be determined.
To further simplify notation let $\vS(u) = \tilde{\vB}'(u) \tilde{\vB}'(u)^t$.
Using the definition of divergence and the equality $\textmd{div}(\vc^t \vS(u) \vc \ \nabla u) = \vc^t \vS'(u) \vc \nabla u \ (\nabla u)^t + \vc^t \vS(u) \vc \ \Delta u$ the variation of the energy in the direction of $v$ can be formulated as
 \begin{equation*}
R_u v = -  \brae{ \textmd{div}\brar{\vc^t \vS(u) \vc \ \nabla u} + \vc^t\vS(u) \vc \ \Delta u  } v \enspace .
\label{eq:7}
\end{equation*}
With the derived functional derivative we are able to obtain the PDE
\begin{equation}
\left\{\begin{array}{rcl}
   \partial_t u & = & \textmd{div}\brar{ \vc^t\vS(u) \vc \ \nabla u} + \vc^t\vS(u) \vc \ \Delta u   \\
   \bras{\nabla u , \vn} & = & 0 ,
\end{array}\right. 
\label{updscheme} 
\end{equation}
where $\lambda>0$ is a constant. The matrix $\vS(u)$ is a symmetric matrix with entries as a block of size three centered around the main diagonal (cmp.\ Figure \ref{fig:StructureS}a). Since $\vS(u)$ is the outer product of the vector $\tilde{\vB}'$, it is positive semi definite. The scalar value $\vc^t\vS(u) \vc$ acts as a weight and we obtain a coefficient with large entires in homogeneous areas as the box decoding includes almost all relevant values. At edges the coefficient has low entries. At outlier positions the coefficient is still large (cmp.\ Figure \ref{fig:StructureS}b).  

We observe that the right hand side of (\ref{updscheme}), top, consists of two  terms, where the left one has the usual form of non-linear diffusion with spatial varying diffusivity $\vc^t\vS(u) \vc$, and the right one the form of a diffusion term ignoring the spatial variation of $\vc^t\vS(u) \vc$. 
For linear decoding channel smoothing breaks down to simple local averaging of $u$ and $\vc^t\vS(u) \vc \equiv1$, independent of the variance of $w$ (cmp.\ (\ref{eq.chanSmooth})), and (\ref{updscheme}) becomes plain linear diffusion (\ref{linear_diffusion}). Using robust decoding $\vc^t\vS(u) \vc$ becomes a spatially varying function and  (\ref{updscheme}) a non-linear diffusion with  reduced diffusivity at edges and high diffusivity in homogenous regions as well as at outlier positions (cmp.\ Figure \ref{fig:StructureS}b).

\subsection{Introducing channel representation to non-linear image diffusion}
Here the LCD will be extended to its non-linear pendant including a convex potential function $\Phi$. The motivation is to further control the filtering to preserve fine image details, similarly to the approach by Perona and Malik~\cite{Perona90scale-spaceand}. Including a potential function allows to suppress outliers further as they will not appear in the structure estimation. 

Let the regularization term be defined as
\begin{equation}
  R(u) = \int_{\Omega} \Phi(|\nabla(\vc^t  \tilde{\vB}(u))|) \,dxdy \enspace ,
\label{eq:}
\end{equation}
where $\vc^t  \tilde{\vB}(u)$ is the channel smoothed version of the image $u$ as defined in (\ref{chansmoothedu}). By computing the variational derivative $R_u v$ in a similar way as presented in the previous section we obtain 
\begin{equation}
\begin{split}
R_u v &= \int_{\Omega} \Phi'(|\nabla(\vc^t  \tilde{\vB}(u))|) \frac{1}{2} \frac{1}{|\nabla(\vc^t  \tilde{\vB}(u))|} \\
&\hspace{5mm}\cdot \frac{\partial}{\partial \epsilon} \brae{|\nabla(\vc^t  \tilde{\vB}(u+\varepsilon v))|^2 }_{\epsilon = 0} \ dxdy .
\label{eq:nonlinenergy}
\end{split}
\end{equation}
Using the short notation $\Psi = \frac{\Phi'(|\nabla(\vc^t  \tilde{\vB}(u))|)}{|\nabla(\vc^t  \tilde{\vB}(u))|}$ and the derivation from the previous section, the EL-equation reads
\begin{equation}
\left\{\begin{array}{rcl}
\partial_t u &=&  \lambda \left[  \dfrac{1}{2} \Psi \right. \ \dive \left (\vc^t\vS(u)\vc\right) \nabla u \\
          &&\hspace{5mm} +\vc^t\vS(u)\vc \ \dive  (\Psi \nabla u )\bigg] \\
 \bras{\nabla u, \vn} & = & 0 ,
\end{array}\right.
\label{eq:updNLCD}
\end{equation}
where $\lambda>0$ and $\vS(u) = \tilde{\vB}'(u) \tilde{\vB}'(u)^t$ as before.\\
The two right hand terms of (\ref{eq:updNLCD}) both implement non-linear diffusion weighted by an additional factor. In the first term the diffusivity stemms from the channel representation and the usual edge stopping function $\Psi$ acts as additional weight. In the second term the roles of $\Psi$ and $\vc^t\vS(u)\vc$ are exchanged. Furthermore, $\Psi$ is defined on the channel smoothed image $\vc^t  \tilde{\vB}(u)$, so outliers are not present in the structure estimation of $\Psi$ and will be smoothed strongly.

\section{Experiments}
In this section we evaluate the proposed linear channel diffusion (LCD) and the non-linear channel diffusion (NLCD) for different grey-scale images. Standard images as "Cameraman" are used as well as images from the Berkeley image database \cite{berk} commonly used for segmentation purposes. Since we are interested in investigating the case of a mixture of noise models we corrupt the images with Gaussian noise as well as impulse noise. Here we consider the presence of  $5\% $ impulse noise and vary the standard deviation of Gaussian noise $\sigma \in \left\{5, 10, 15, 20, 30, 40, 50\right\}$. 

\subsection{Setup of evaluation}

\renewcommand{\xlev}{150}
\renewcommand{\xlevtitle}{212}
\renewcommand{\imWidth}{.30}

\begin{figure*}[t]
\centering
\begin{picture}(300,340)
  \put(-45,0){\includegraphics[width=.20\textwidth]{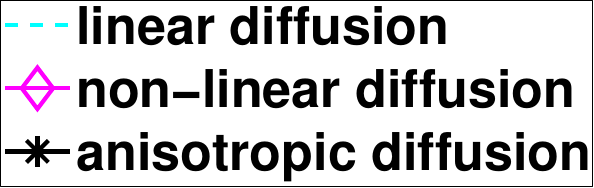}}
  \put(100,2){\includegraphics[width=.25\textwidth]{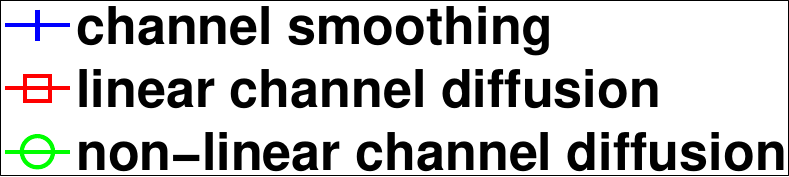}}
  \put(280,1){\includegraphics[width=.16\textwidth]{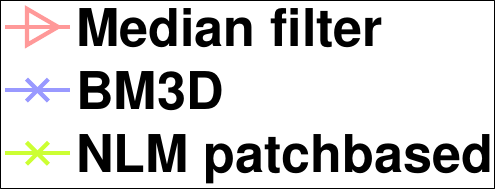}}

  \put(-80,250){ \includegraphics[width=\imWidth\textwidth]{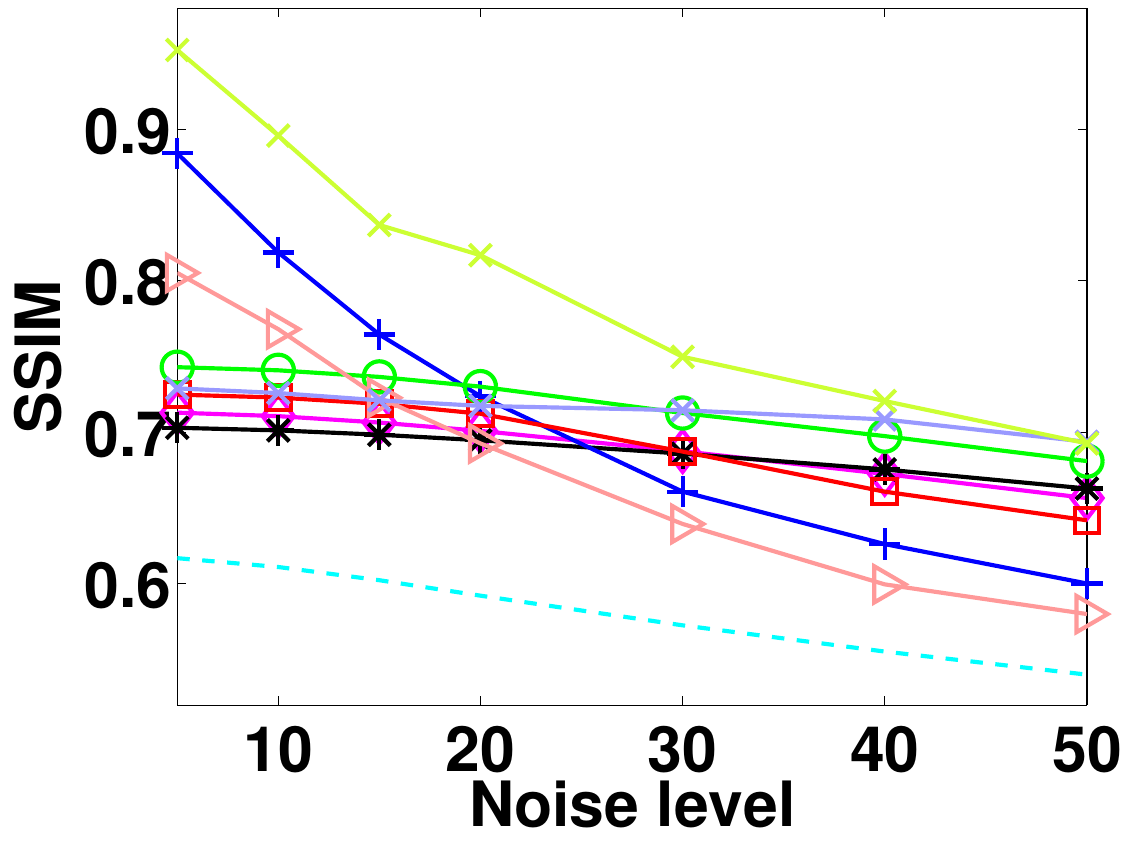}}
  \put(-10,340){\textbf{Cameraman}}

  \put( 80,250){ \includegraphics[width=\imWidth\textwidth]{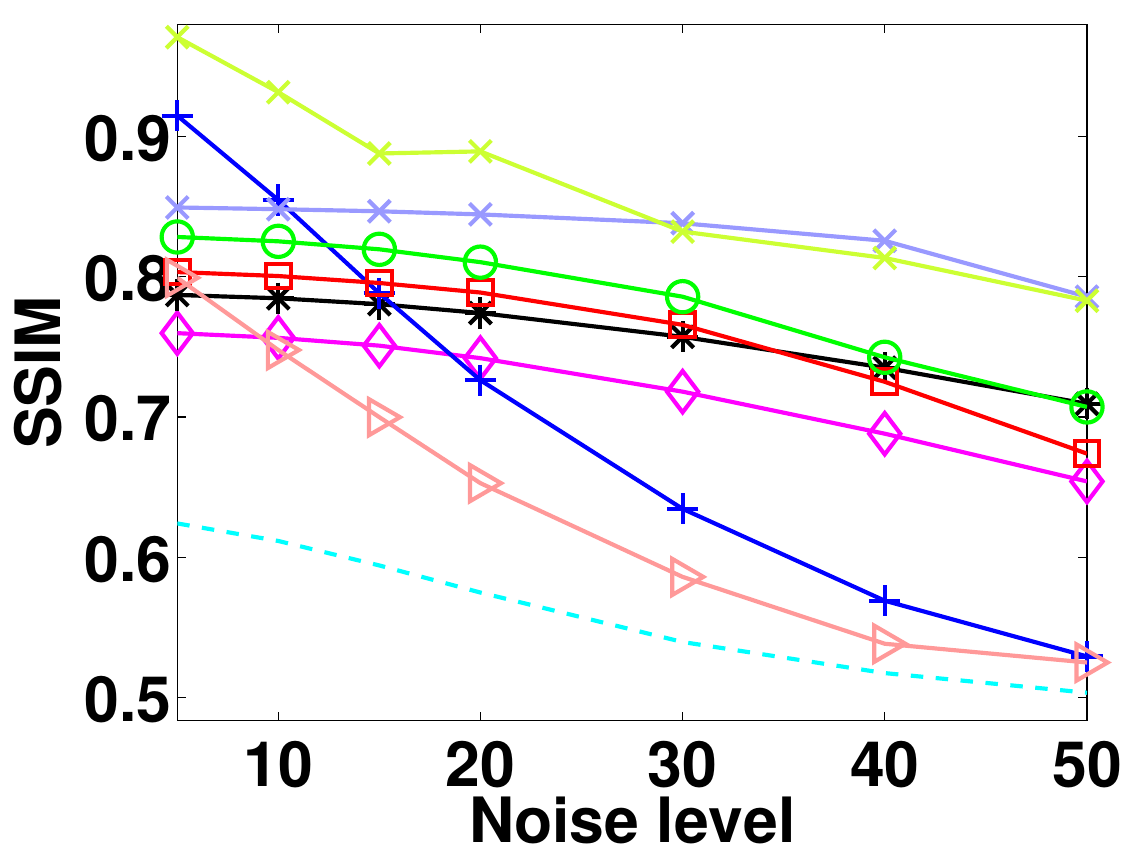}}
  \put(150,340){\textbf{Skyline}}
  
  \put(240,250){ \includegraphics[width=\imWidth\textwidth]{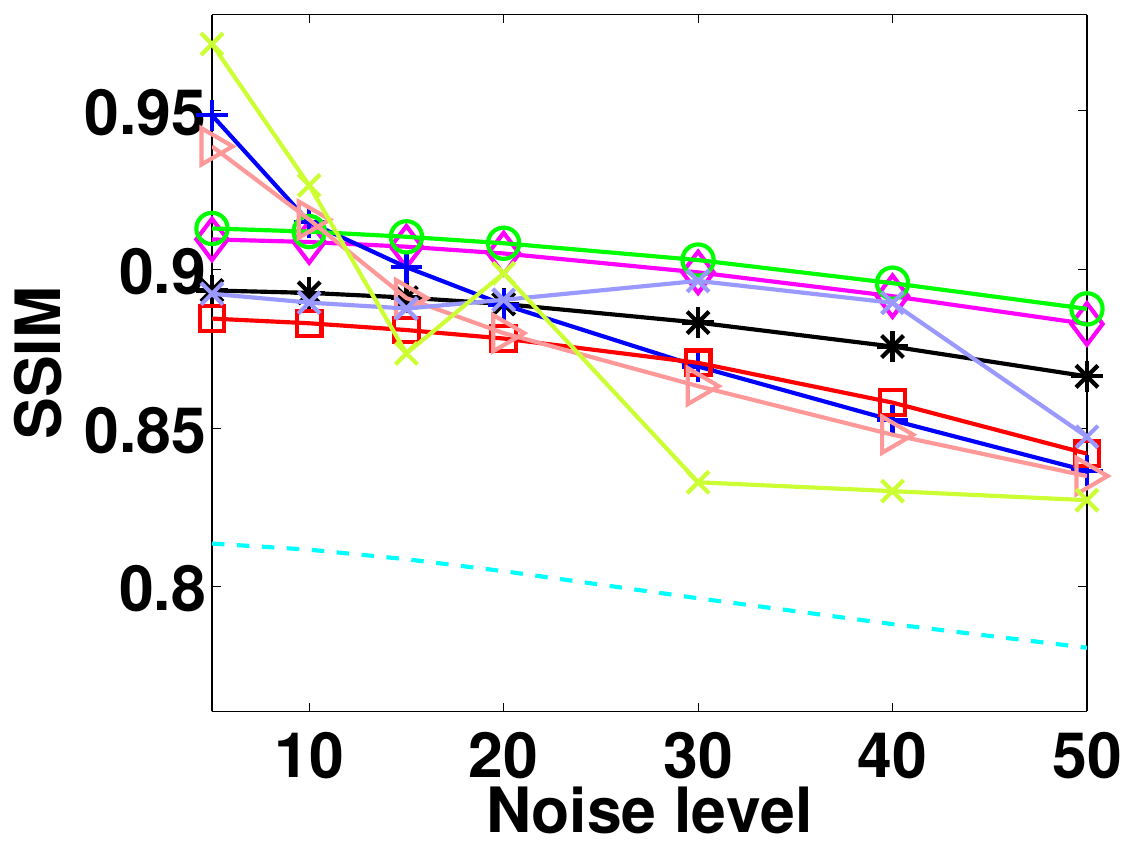}}
  \put(310,340){\textbf{Hawk}}
  
  \put(-80,145){ \includegraphics[width=\imWidth\textwidth]{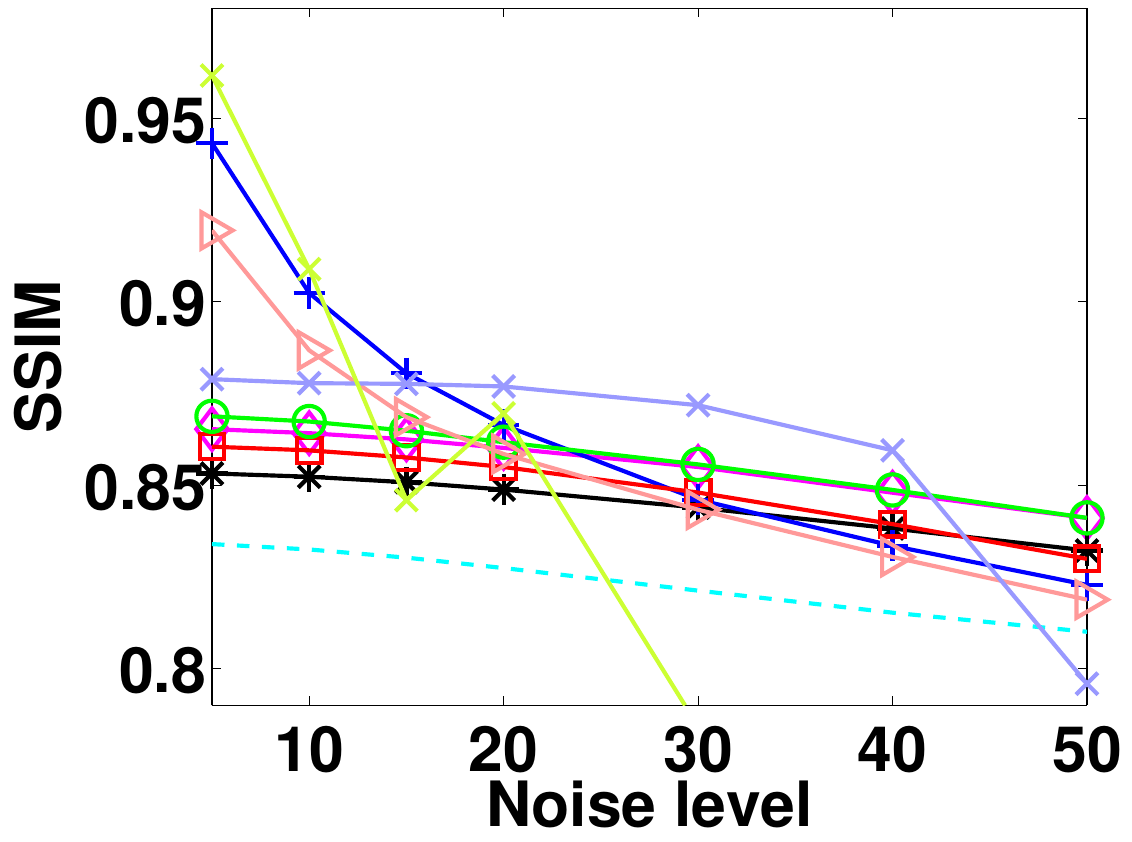}}
  \put(-10,235){\textbf{Bird}}
  
  \put( 80,145){ \includegraphics[width=\imWidth\textwidth]{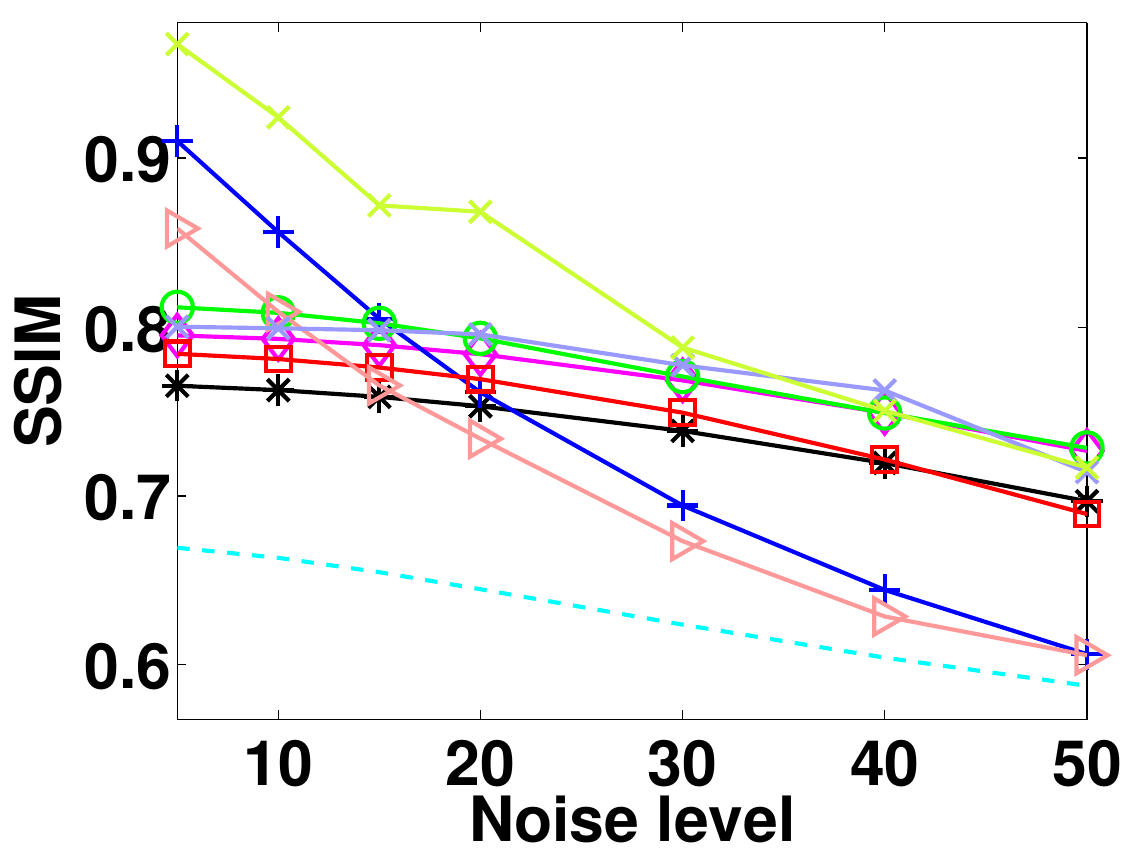}}
  \put(150,235){\textbf{Rollercoaster}}
  
  \put(240,145){ \includegraphics[width=\imWidth\textwidth]{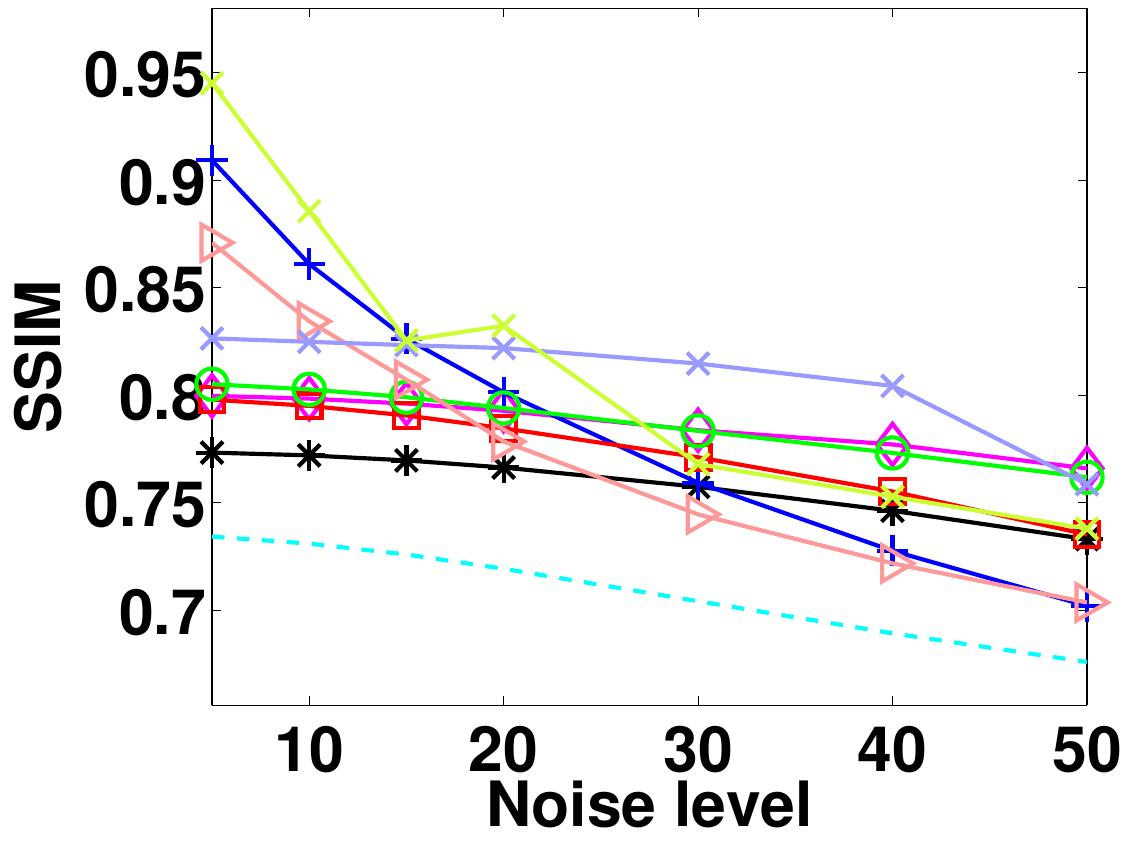}}
  \put(310,235){\textbf{House}}
  
  \put(-80,40){ \includegraphics[width=\imWidth\textwidth]{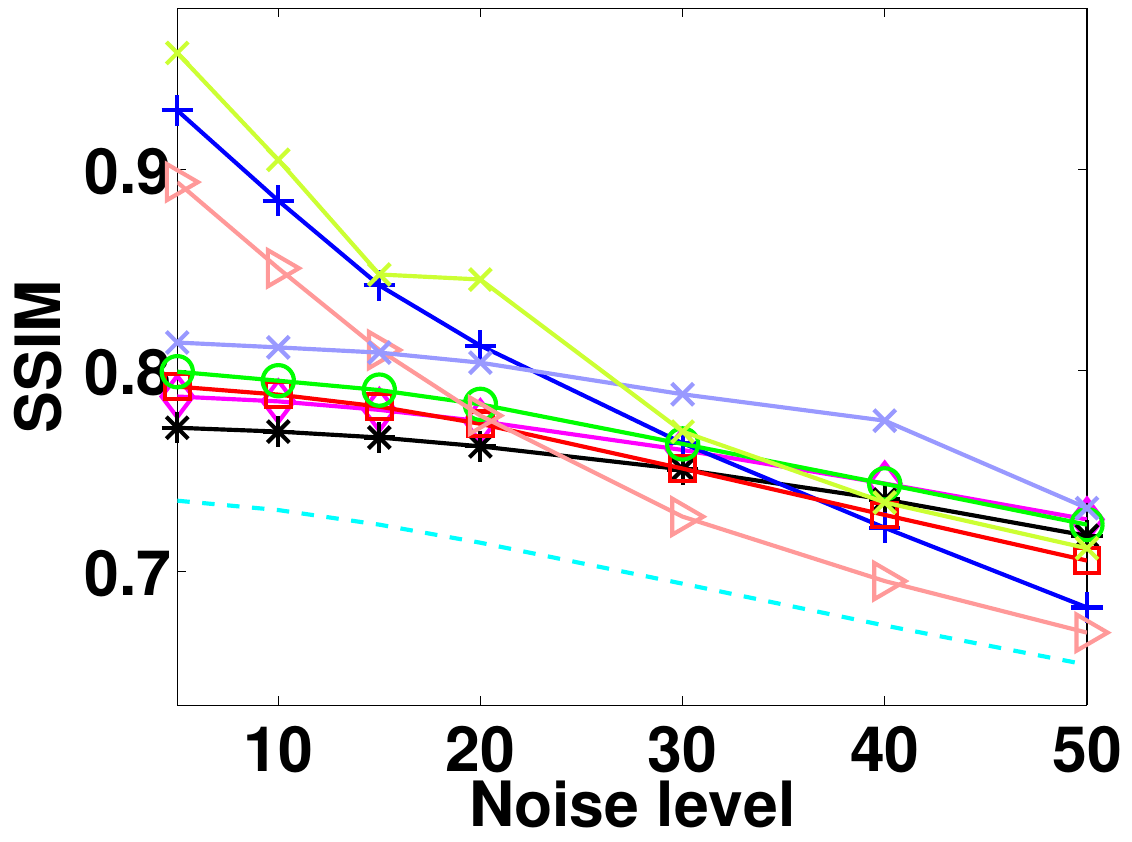}}
  \put(-10,130){\textbf{Lena}}
  
  \put( 80, 40){\includegraphics[width=\imWidth\textwidth]{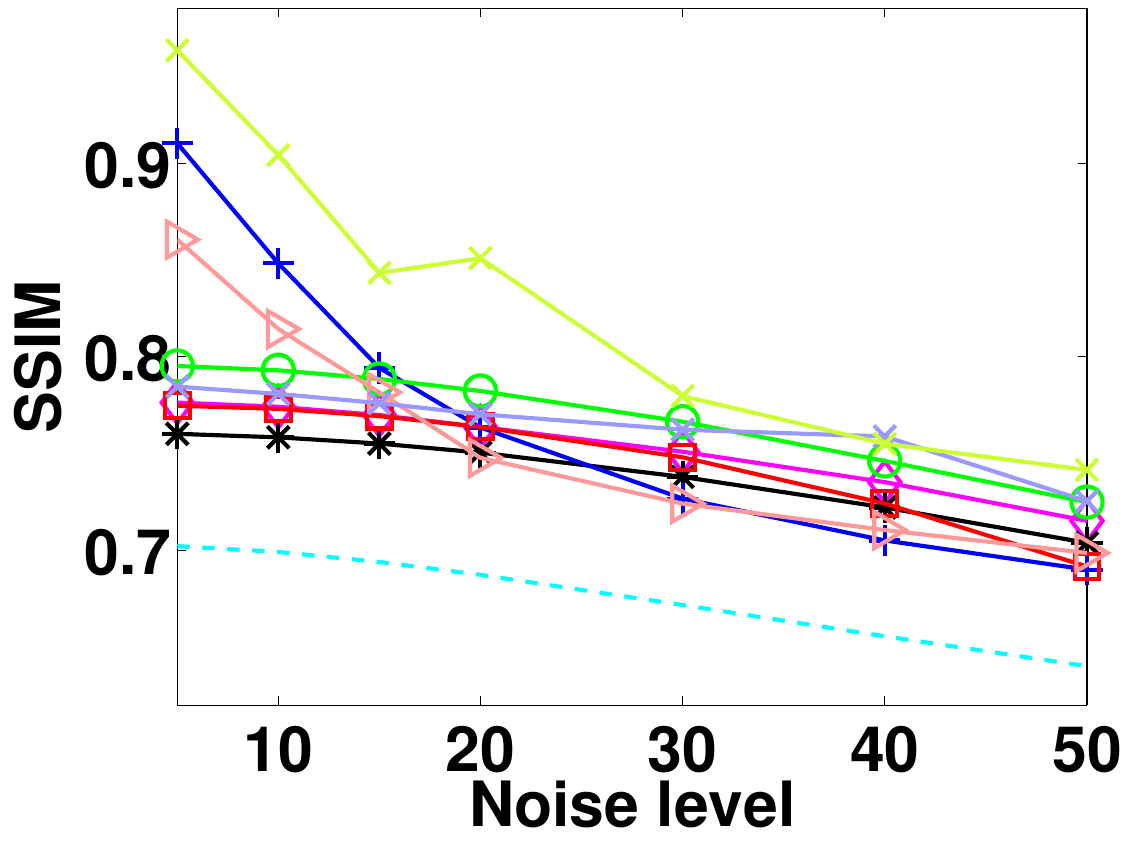}}
  \put(150,130){\textbf{Boat\&House}}
  
  \put(240, 40){\includegraphics[width=\imWidth\textwidth]{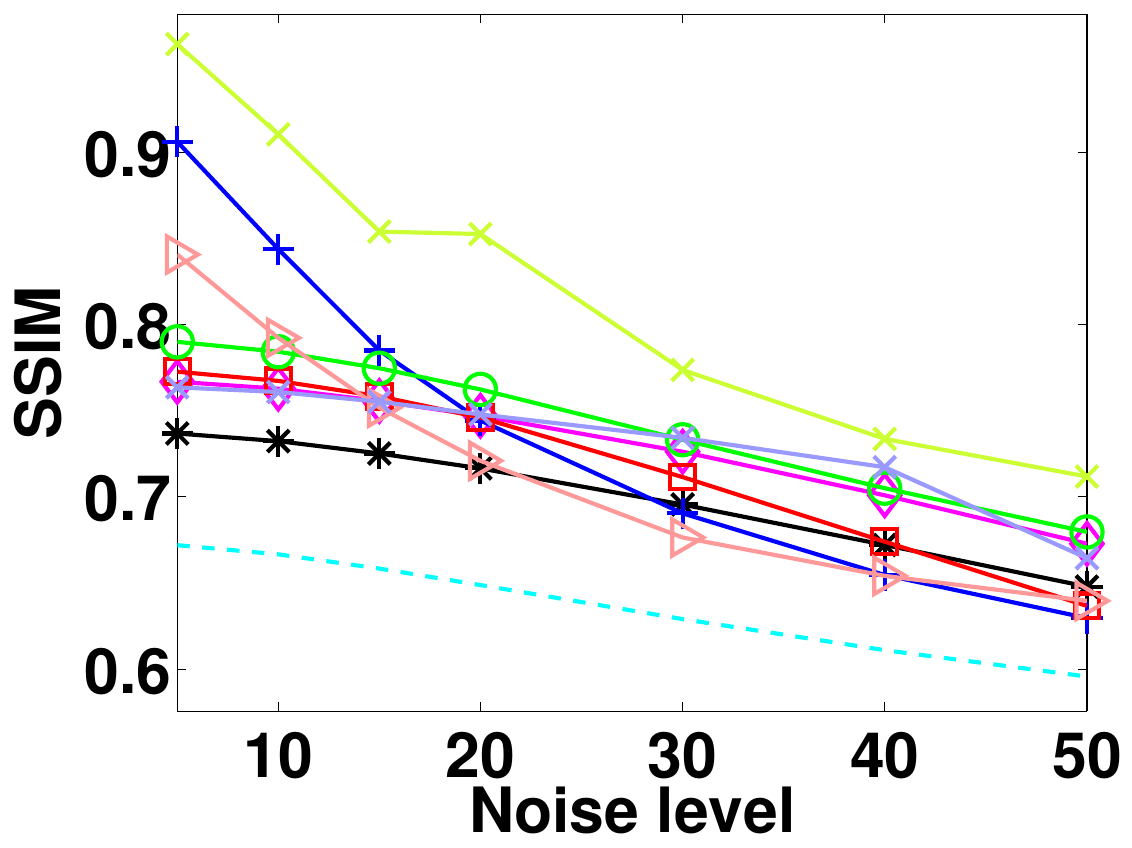}}
  \put(310,130){\textbf{Lighthouse}}
\end{picture}
\caption{Quantitative denoising results of the images Skyline (69007), Hawk (42040), Bird (197027), Rollercoaster (235098), Boat \&House (140088) and Lighthouse (228076) from the Berkeley image database \cite{berk} and standard test images (Cameraman, House, Lena). The numeric name is the same
as in the database. Used methods are linear diffusion, NLD \cite{Perona90scale-spaceand}, AD \cite{Weickert96anisotropicdiffusion}, CS \cite{ffs05}, the novel LCD, the novel NLCD, MF \cite{GonzalezWoods}, BM3D \cite{Dabov06imagedenoising} and NLM \cite{NLMpatch}. SSIM versus different Gaussian noise levels is plotted. For details see text. Best viewed in color}
\label{fig:denoising}
\end{figure*}

The aim of the evaluation is to compare the proposed LCD and NLCD schemes especially to diffusion schemes and channel smoothing as we introduced an extension of these methods. Current state-of-the-art denoising methods are included as well. \\
We compare to linear diffusion (LD) and non-linear diffusion (NLD) as introduced by Perona-Malik \cite{Perona90scale-spaceand}. Furthermore, we consider the tensor-driven anisotropic diffusion scheme (AD) \cite{Weickert96anisotropicdiffusion}. Besides  channel smoothing (CS) \cite{ffs05}, median filtering (MF) is included as a method well suited for impulse noise. Further BM3D \cite{Dabov06imagedenoising} is regarded, currently one of the best methods for filtering Gaussian noise. It considers both a non-linear threshold operation as well as a linear Wiener filter approach to a stack of patches which locally describe the same image region. Finally, two implementations of non-local means are regarded. We compared to the original, pixel based implementation \cite{NLMpixel} as well as to the the patch based variant \cite{NLMpatch}.

All methods have been optimized with respect to their parameters. We optimized LCD and NLCD with respect to the number of channels and the edge parameter $\alpha$ in the edge stopping function $\Psi$, which has been chosen as $\Psi(|\nabla u|) = (1+\frac{|\nabla u|^2}{\alpha^2})^{-1}$. Furthermore, the size of the median filter was optimized. For the channel smoothing (CS) we use a box decoding scheme, optimize the number of channels, and optimize the variance of the Gaussian filter used for smoothing in each channel. In AD, the mapping of the structure tensor eigenvalues to diffusivities is done using the same edge stopping function as for NLD. The parameter $\alpha$ is also optimized.

We implement our diffusions in a standard Euler forward scheme and use finite differences to approximate the image derivatives for every spatial position in the image. For each method and parameter optimization we let the filtering continue until the maximum of the structure similarity index (SSIM) \cite{1284395} has been reached.

\renewcommand{\imgsize}{0.29}
\renewcommand{\imgsizezoom}{0.10}
\begin{figure*}[!ht]
\vfill
\begin{centering}
\begin{tabular}{c ccc ccc}
   %%%%%%%%%%%%%%%%%%%%%%%%%%%%%%%%%%%%%%%%%%%%%%%%%%%%%%%%%%%%%%%%%%%%%%%%%%%%%%%%%%%%%%
   \begin{tabular}{c}     
              Cameraman            \\[0.04cm]                        
     \multicolumn{1}{c}{\includegraphics[width=\imgsize\linewidth]{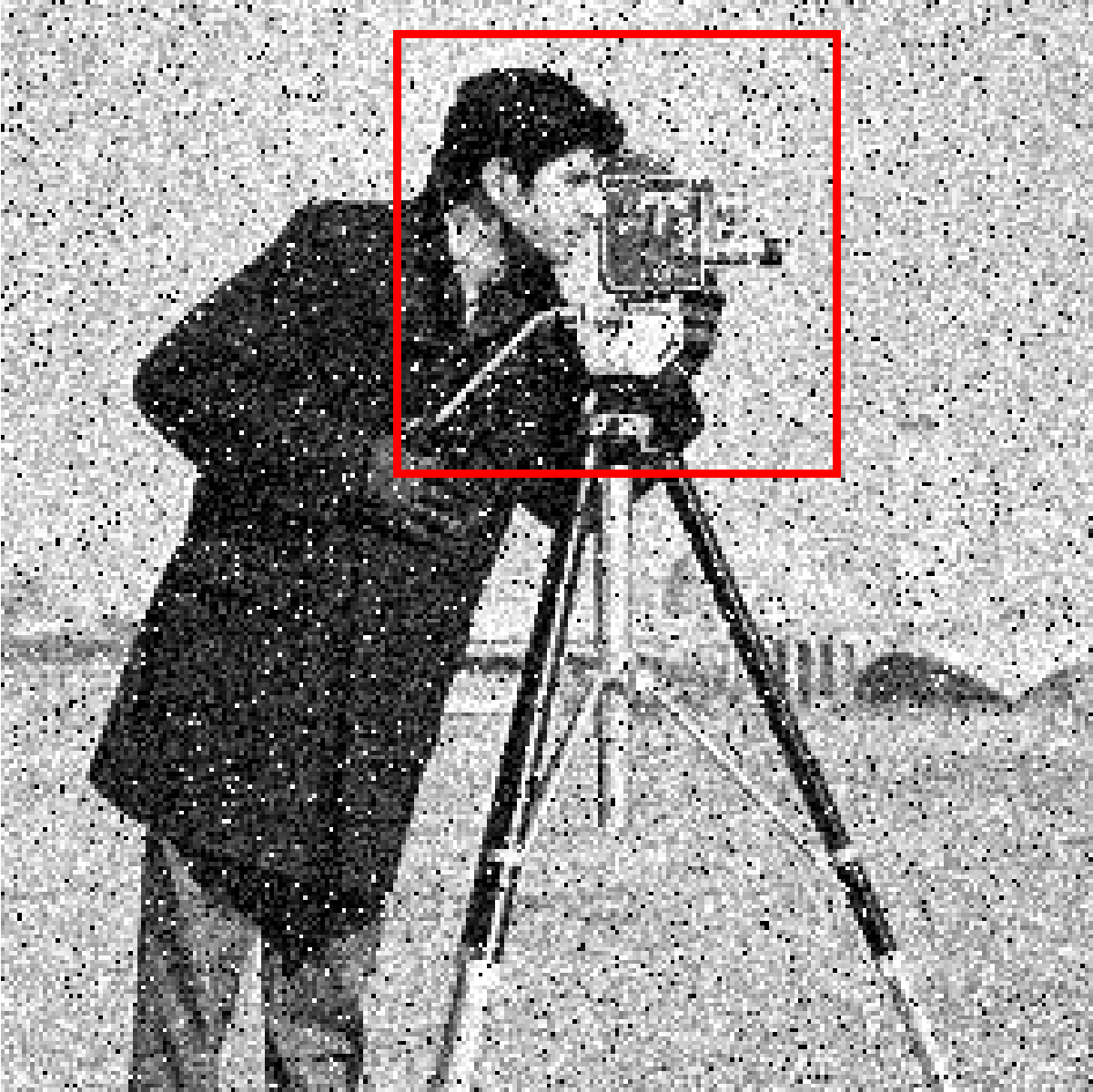}}  \\
     Initial noise = 0.2692 \\[0.7cm]
   \end{tabular}
   % ---------------------
   \begin{tabular}{ccc ccc}                             
      Original &  NLD & AD & NLM pixel & BM3D  \\
    \includegraphics[width=\imgsizezoom\linewidth]{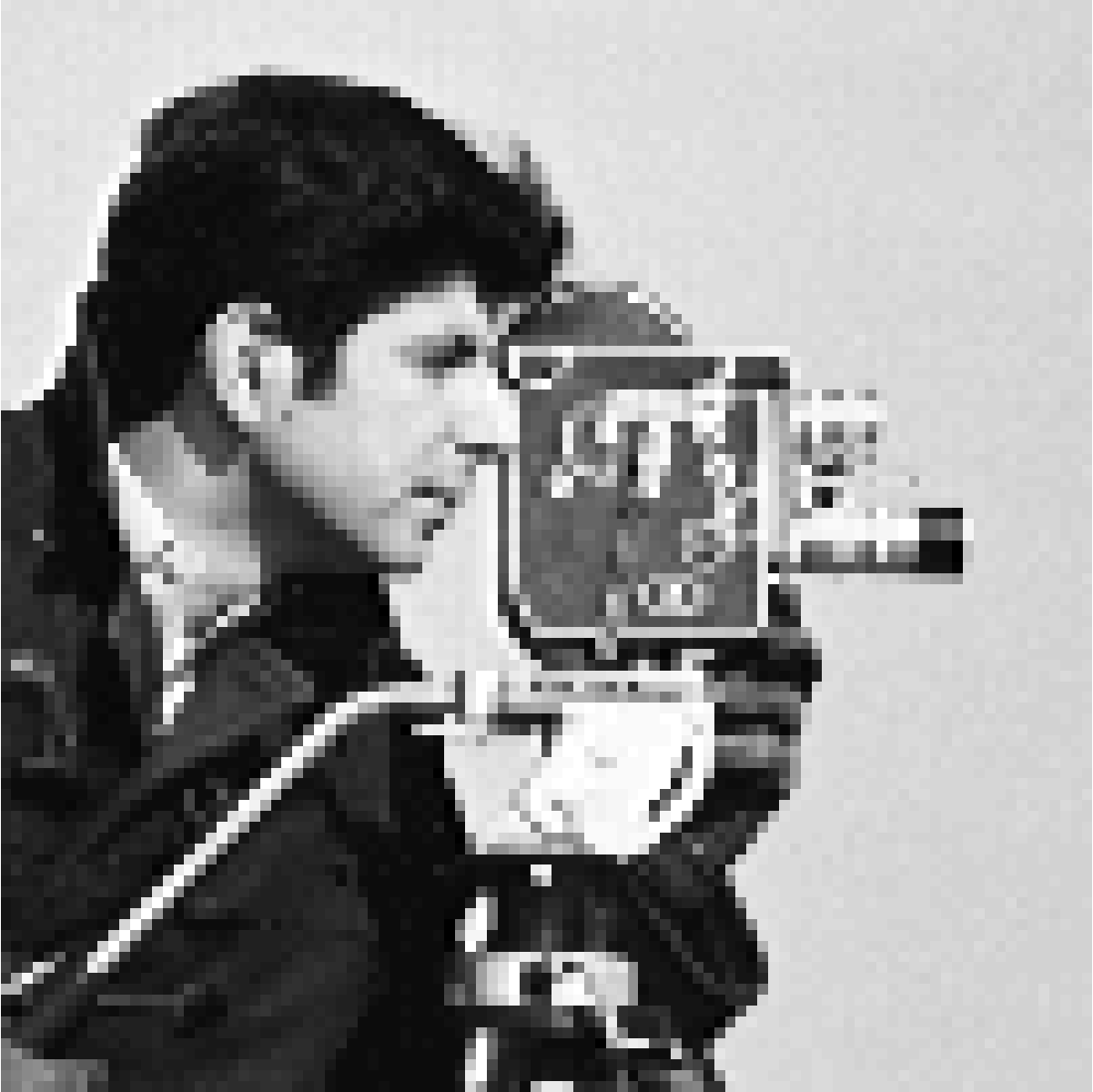}  &
    \includegraphics[width=\imgsizezoom\linewidth]{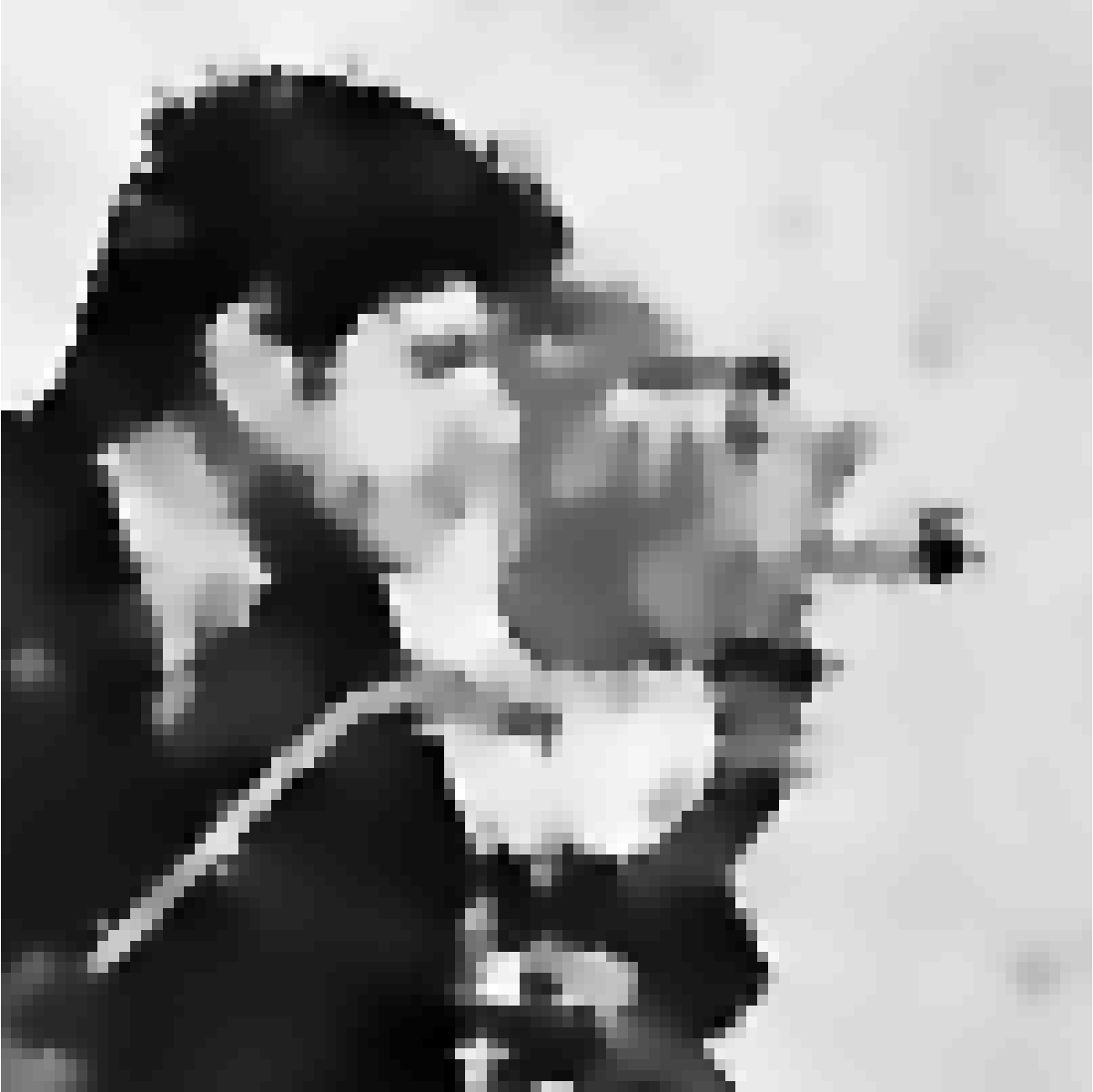}  &
    \includegraphics[width=\imgsizezoom\linewidth]{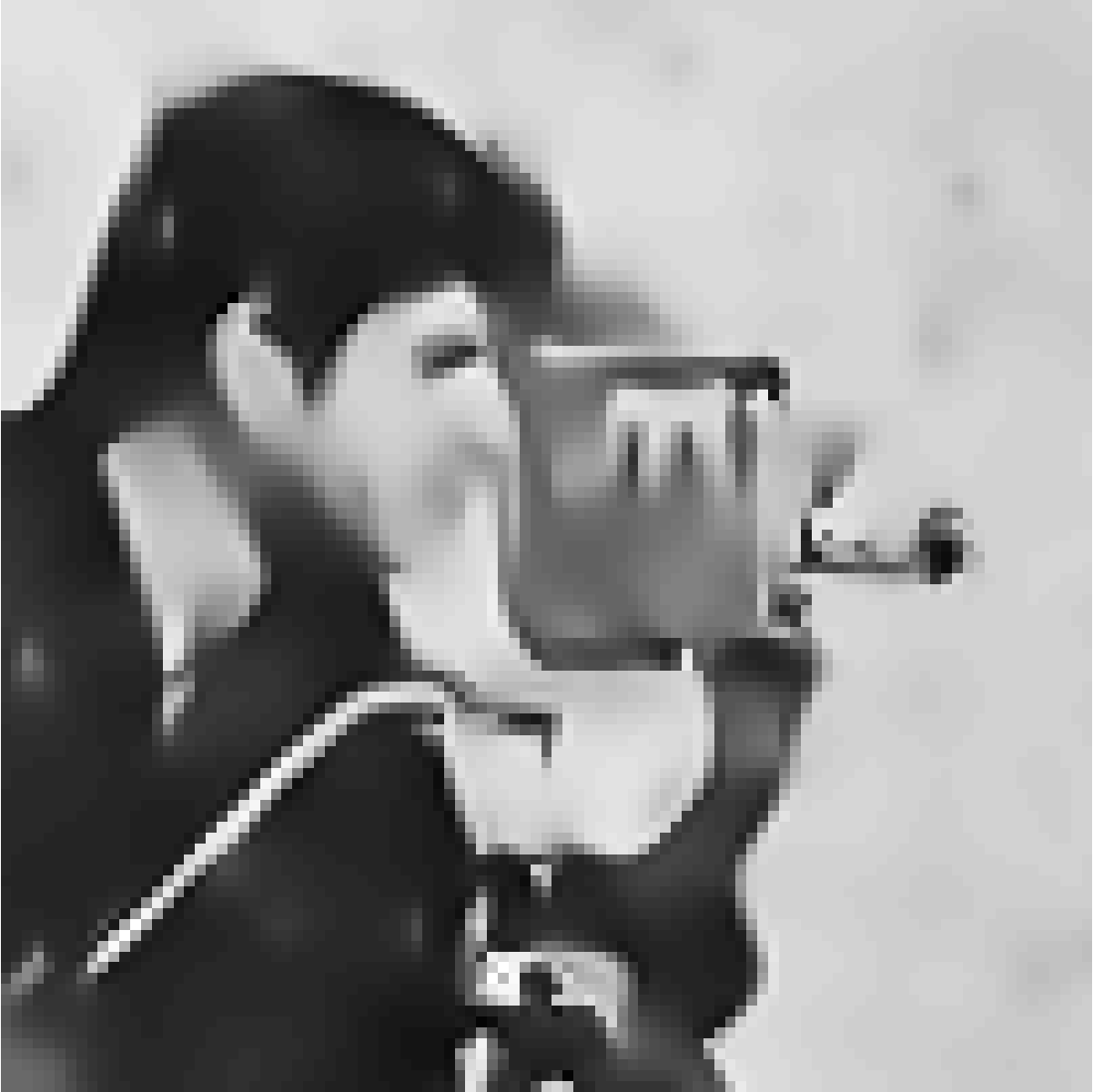}  &
    \includegraphics[width=\imgsizezoom\linewidth]{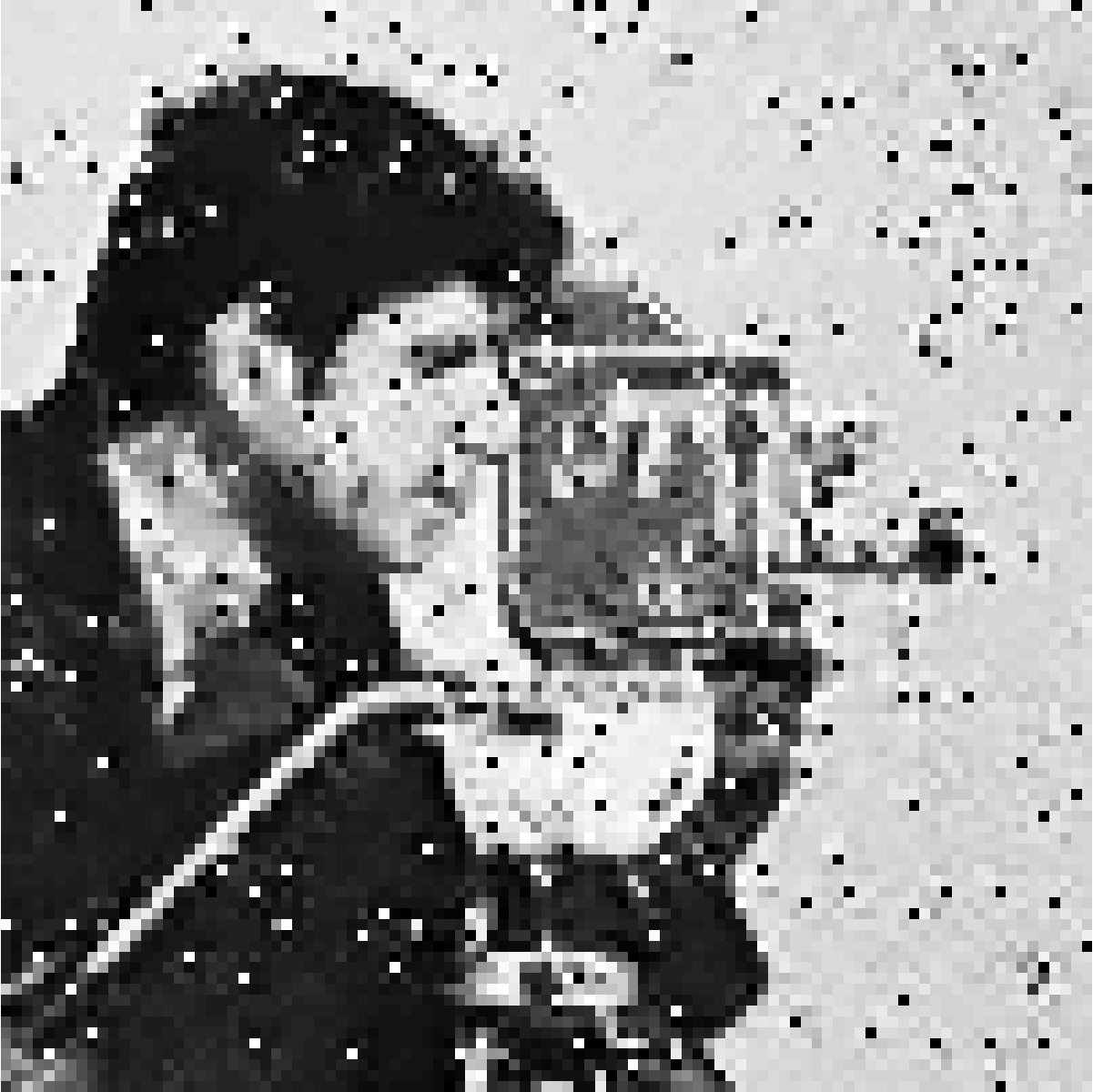} &
		\includegraphics[width=\imgsizezoom\linewidth]{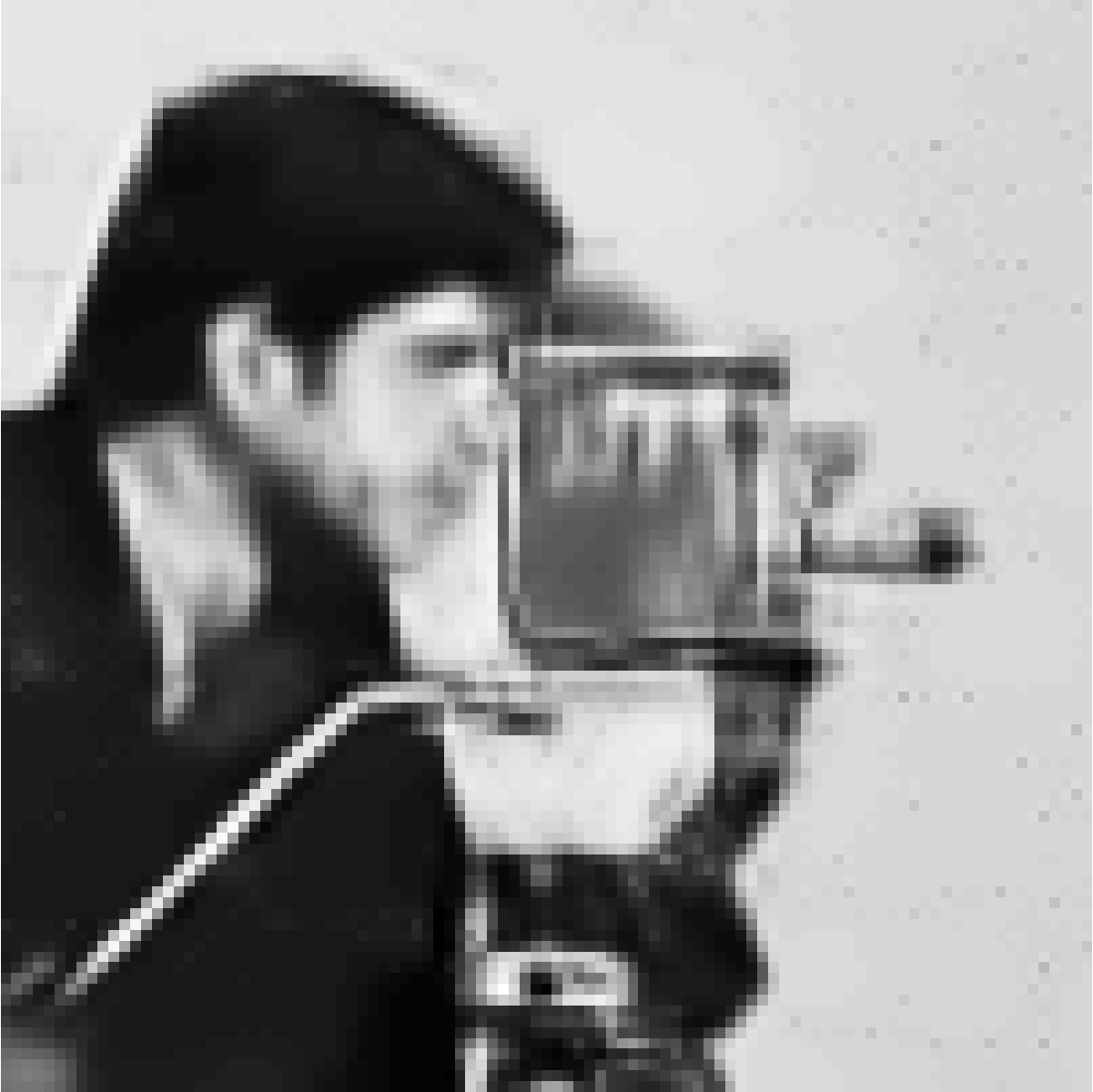} \\
     &  0.6874 & 0.6858 & 0.3753 & 0.7145  \\[0.2cm]
     CS &  MF & NLM patch  & \textit{\textbf{LCD}} & \textbf{\textit{NLCD}}  \\
    \includegraphics[width=\imgsizezoom\linewidth]{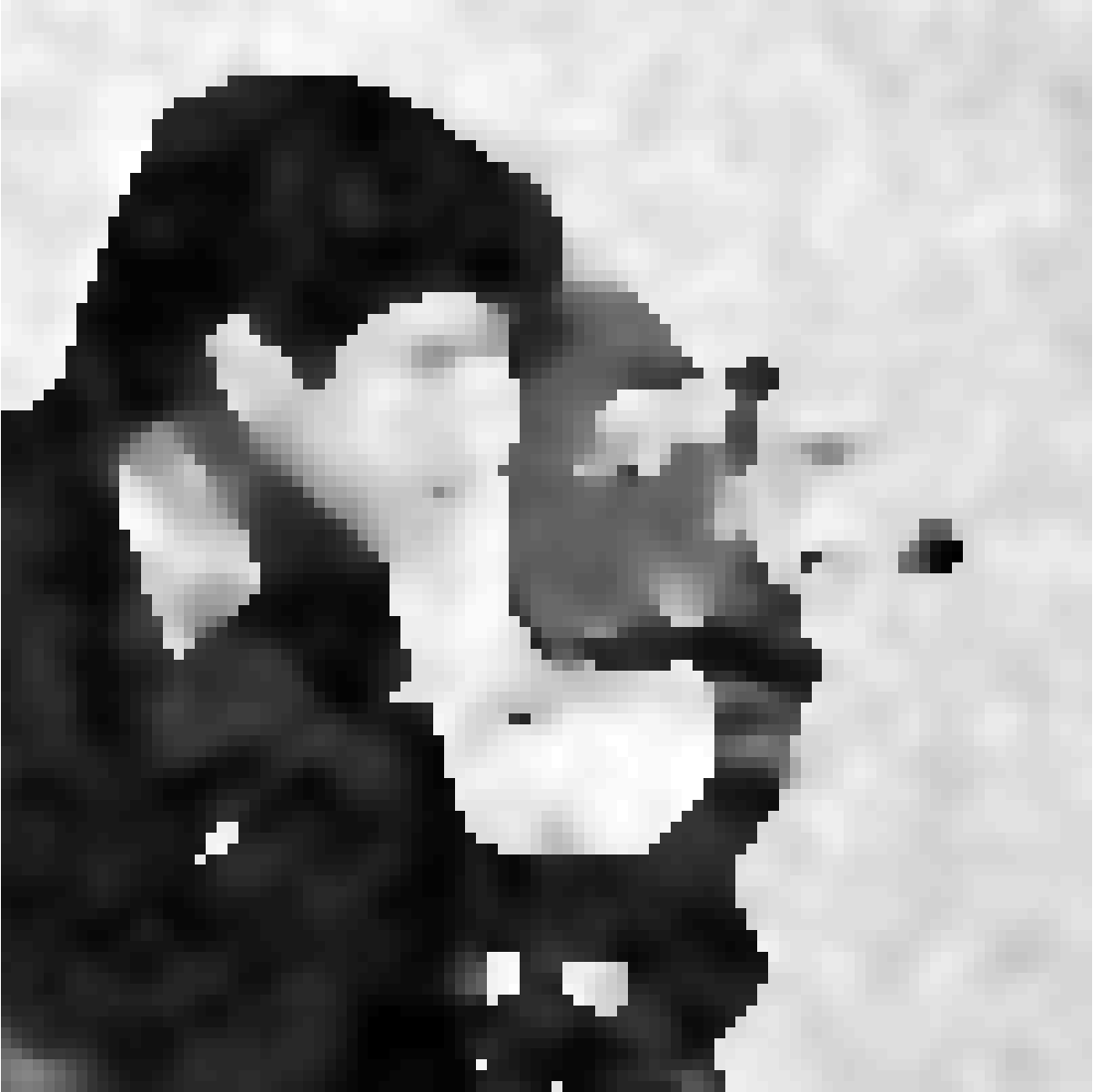}  &
    \includegraphics[width=\imgsizezoom\linewidth]{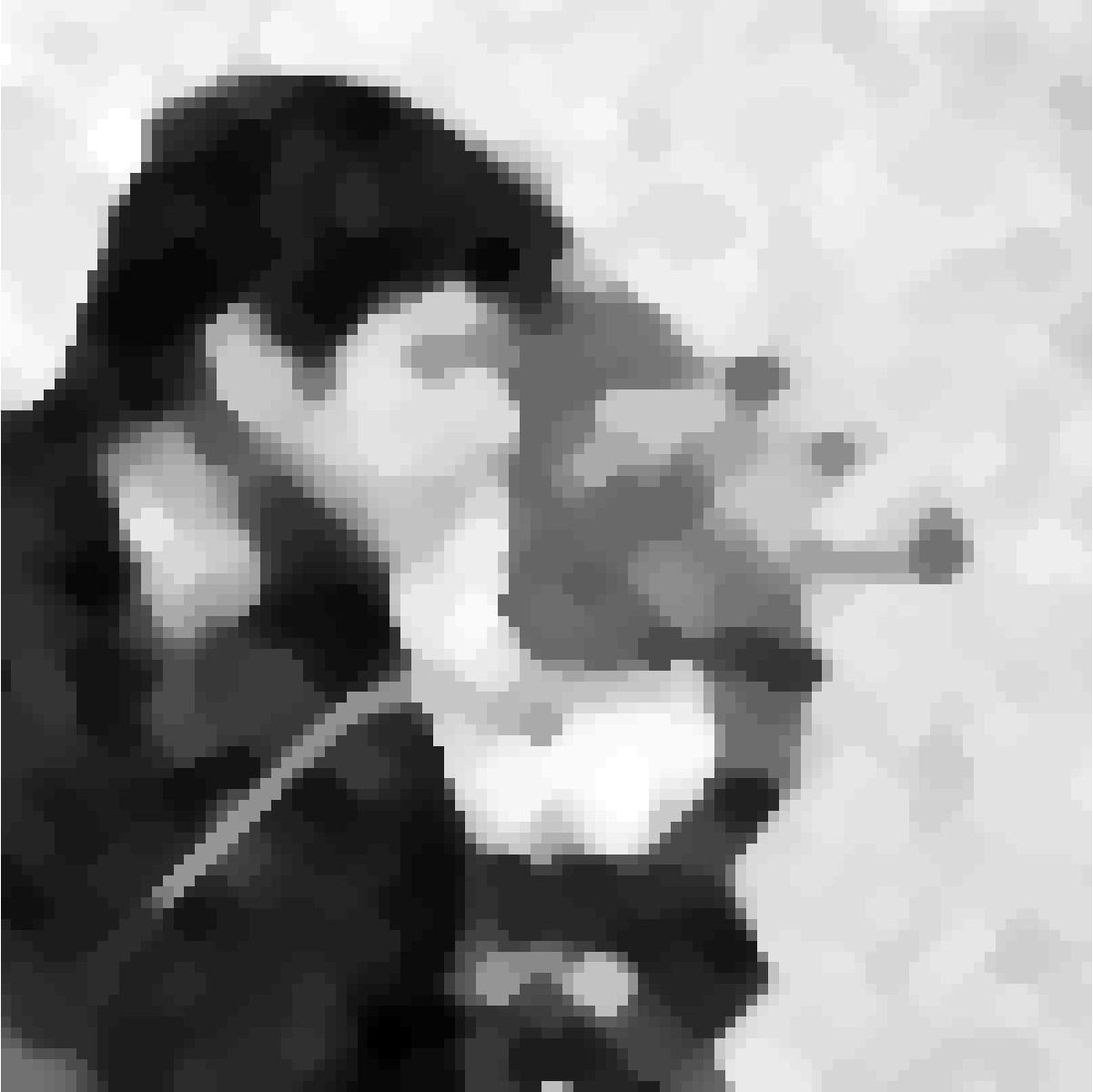} &
    \includegraphics[width=\imgsizezoom\linewidth]{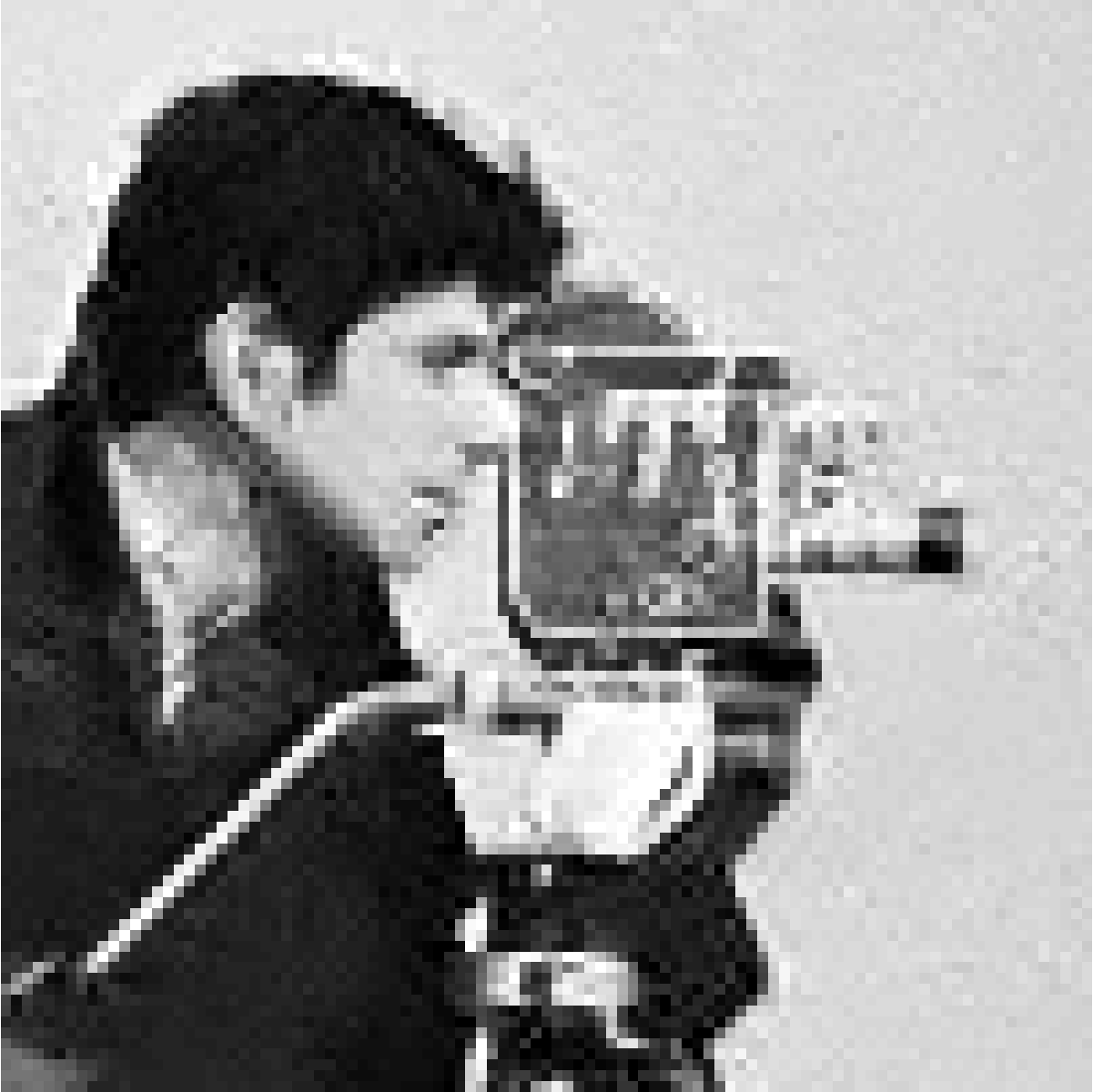}  &
    \includegraphics[width=\imgsizezoom\linewidth]{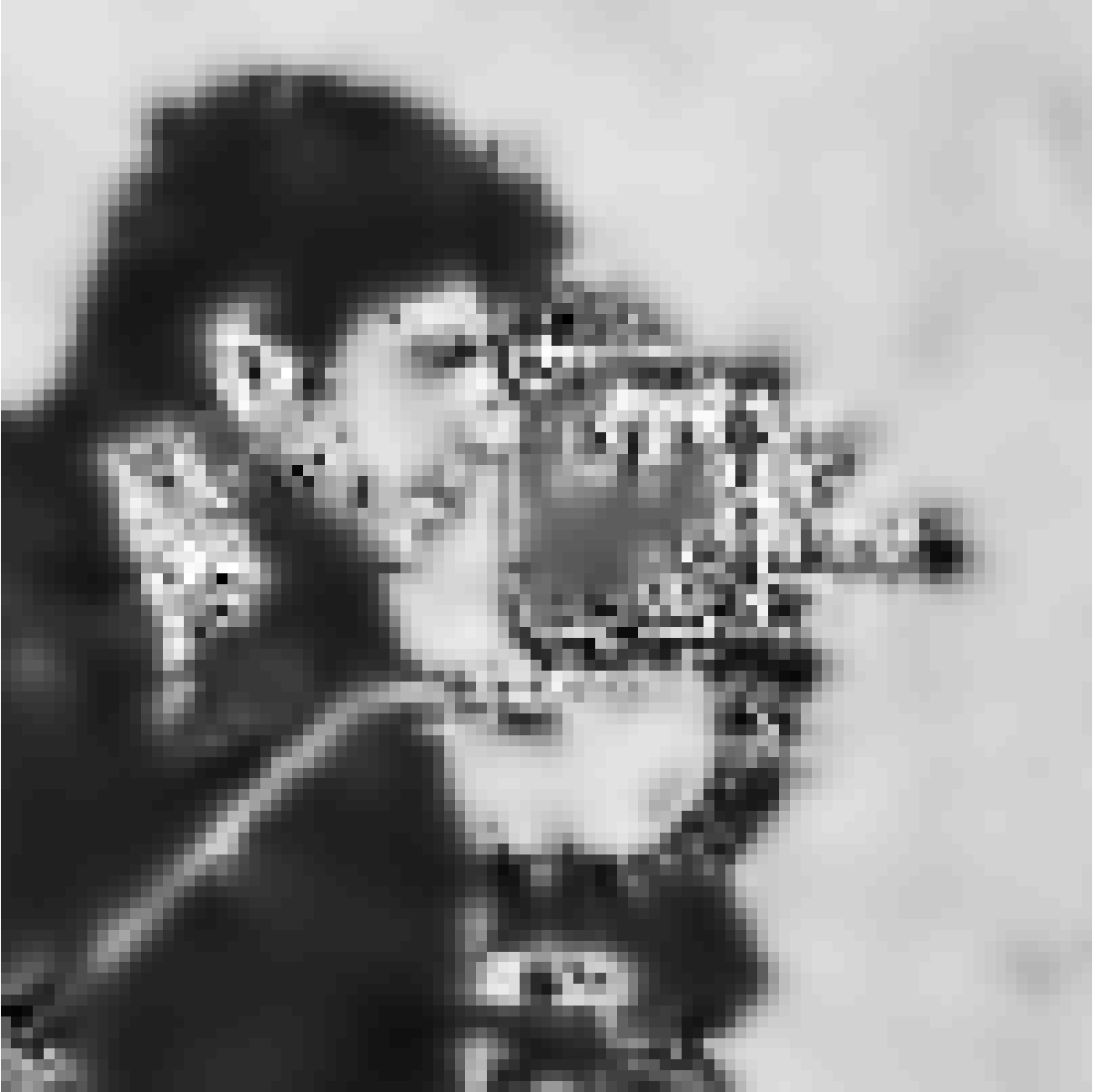} &
		\includegraphics[width=\imgsizezoom\linewidth]{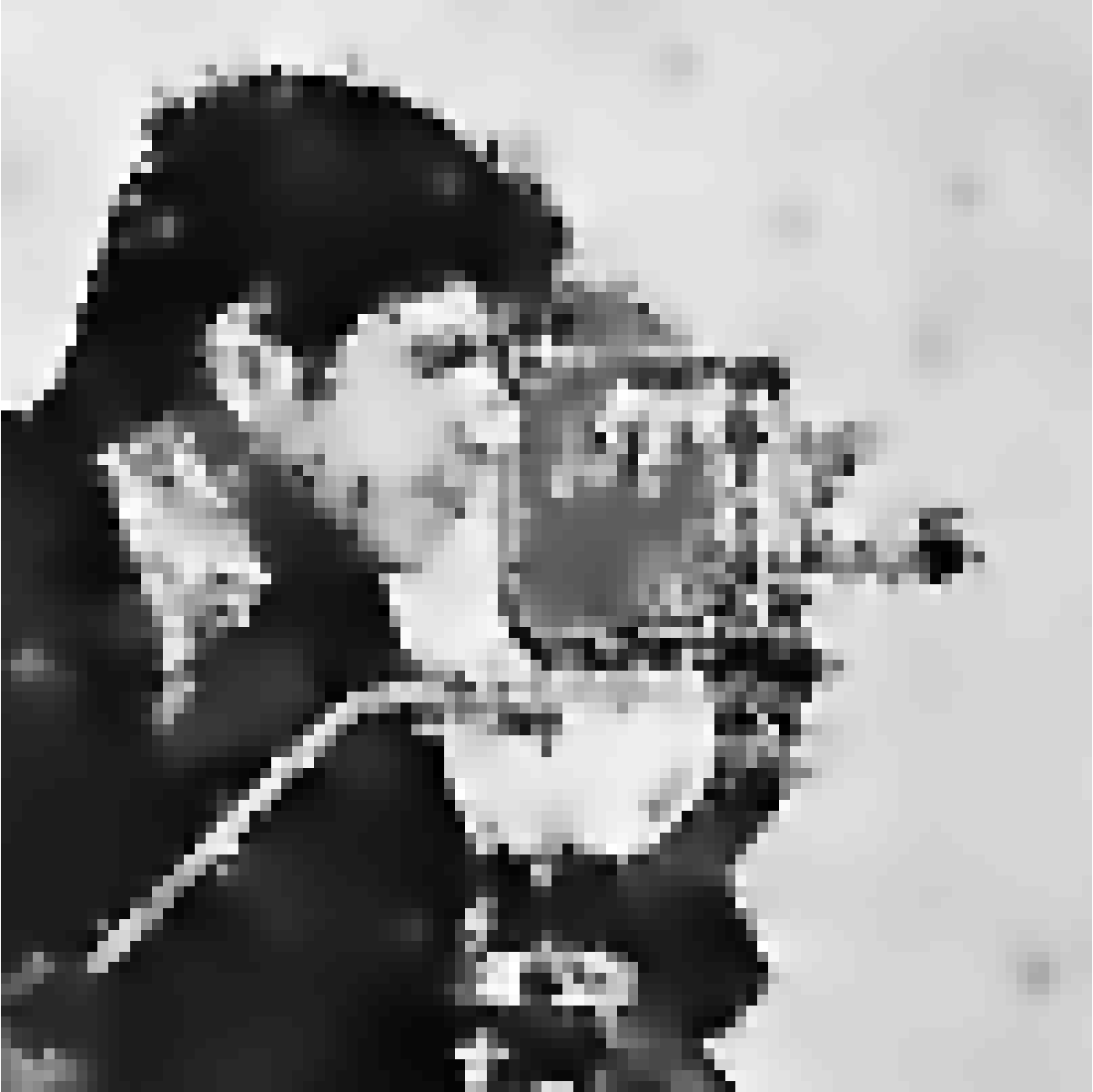} \\
    0.6606 &  0.6393  & 0.7408 & 0.6871 & 0.7126 \\[0.5cm]   \end{tabular} \\[0.1cm]
   %%%%%%%%%%%%%%%%%%%%%%%%%%%%%%%%%%%%%%%%%%%%%%%%%%%%%%%%%%%%%%%%%%%%%%%%%%%%%%%%%%%%%%
   \begin{tabular}{c}     
                 Hawk         \\[0.12cm]                        
     \multicolumn{1}{c}{\includegraphics[width=\imgsize\linewidth]{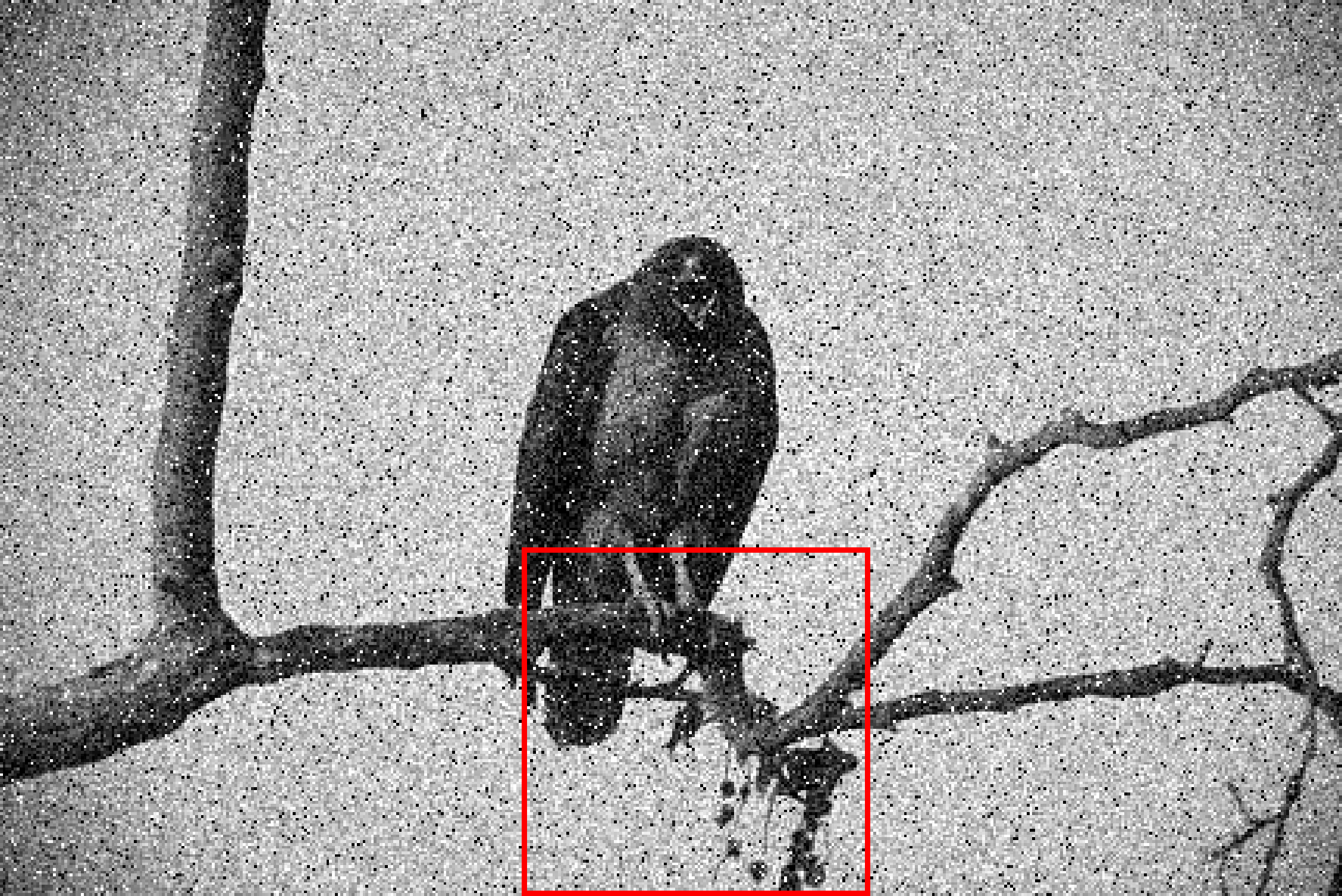}}  \\
     Initial noise = 0.1460 \\[0.1cm]
   \end{tabular}
   % ---------------------
   \begin{tabular}{ccc ccc}                             
      Original &  NLD & AD & NLM pixel & BM3D  \\
    \includegraphics[width=\imgsizezoom\linewidth]{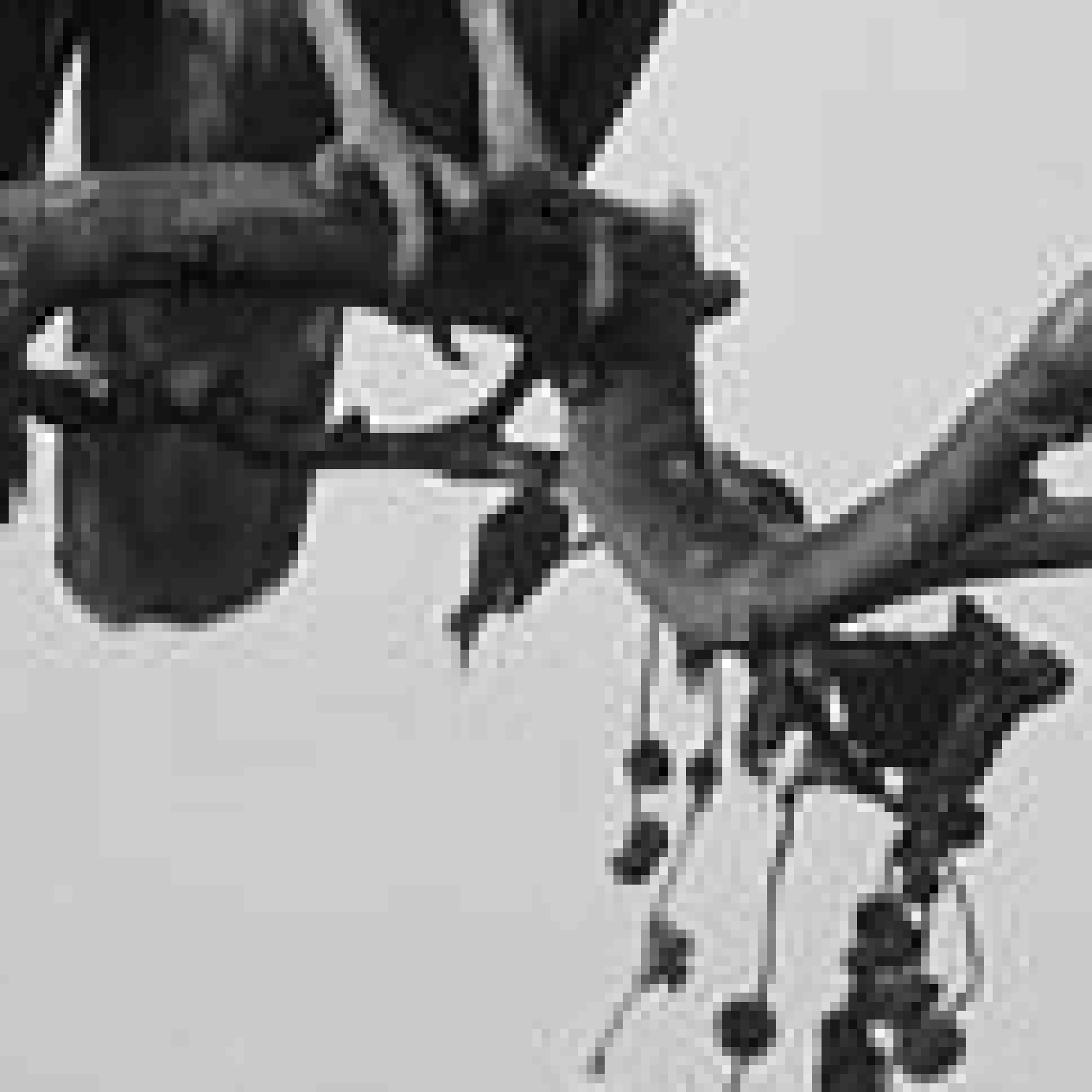}  &
    \includegraphics[width=\imgsizezoom\linewidth]{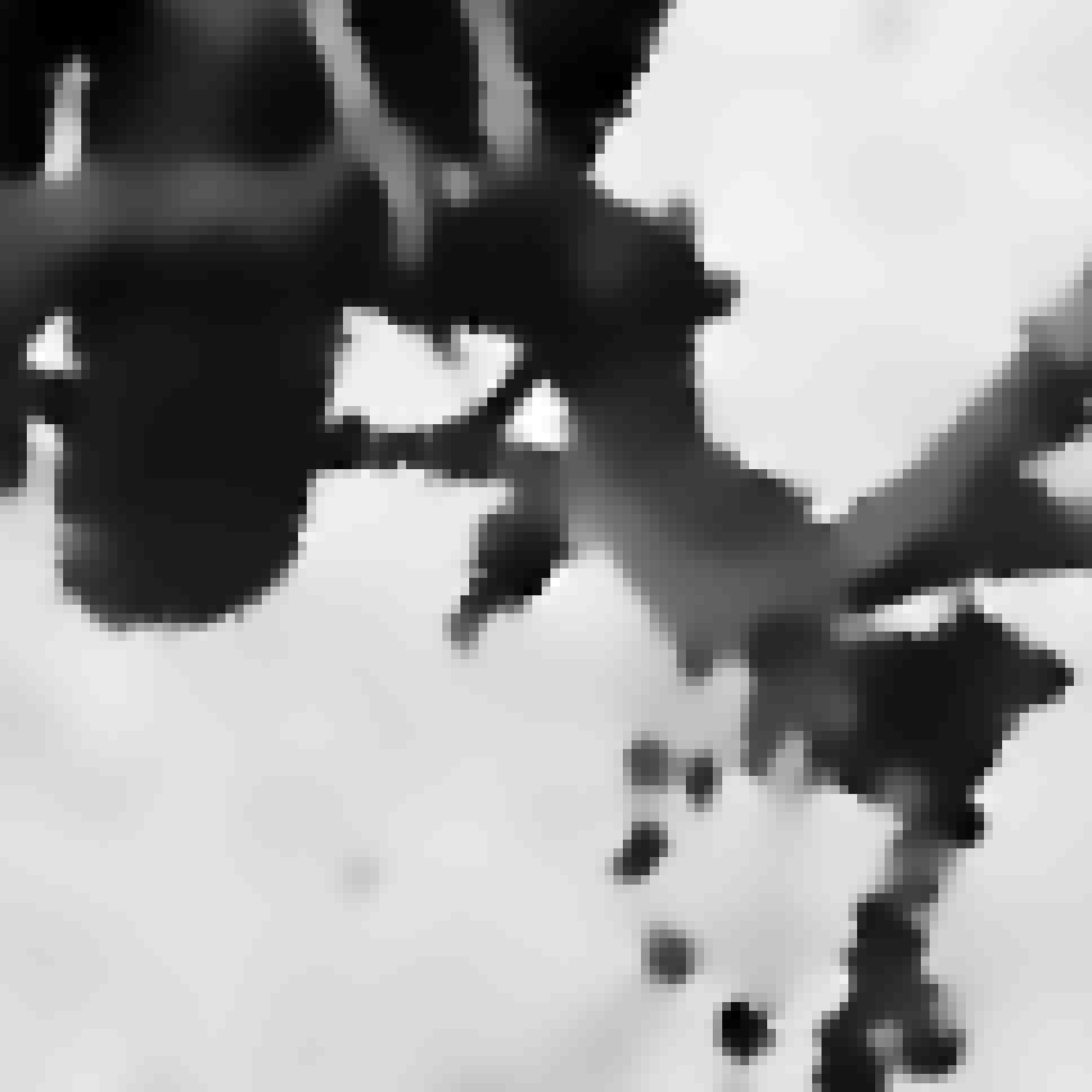}  &
    \includegraphics[width=\imgsizezoom\linewidth]{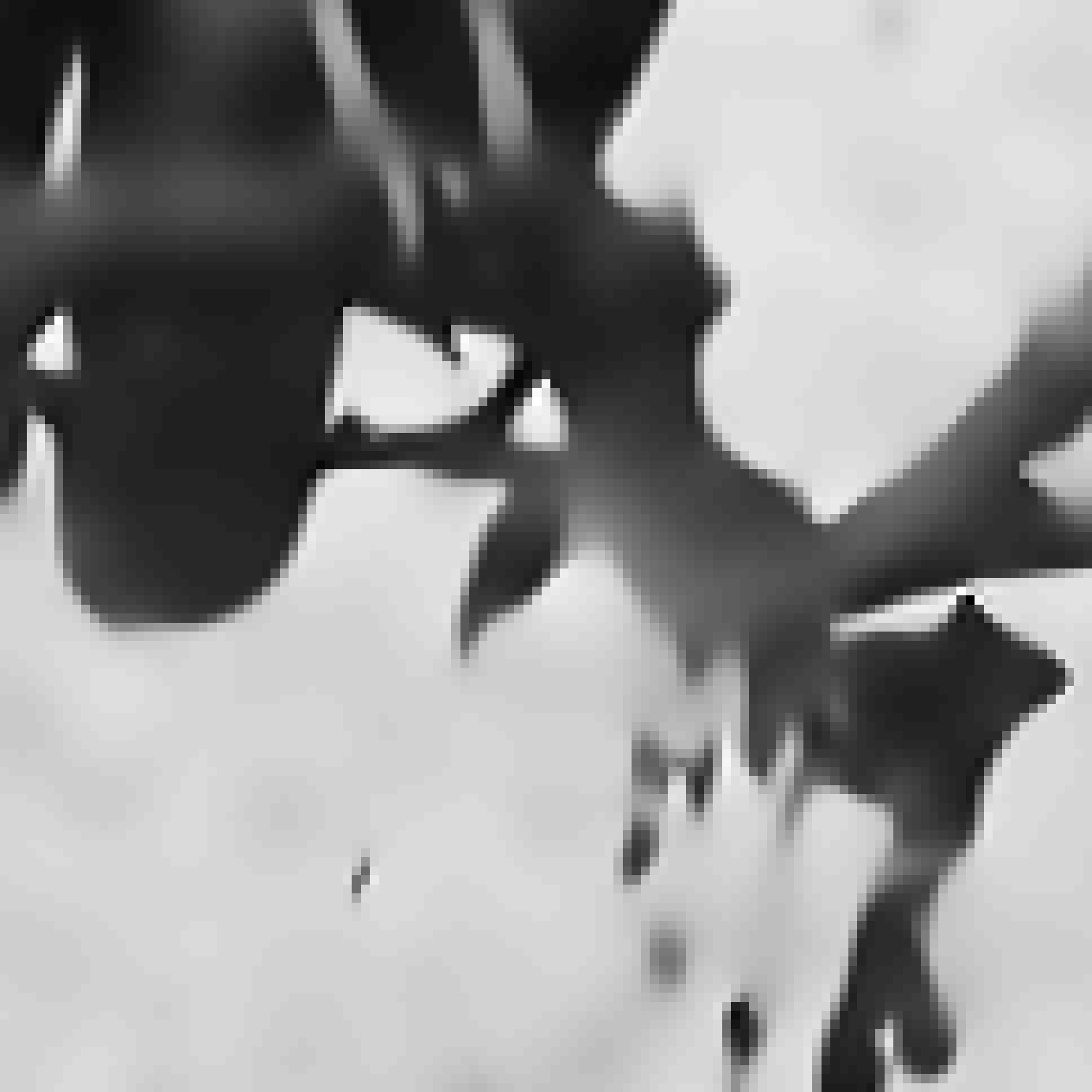}  &
    \includegraphics[width=\imgsizezoom\linewidth]{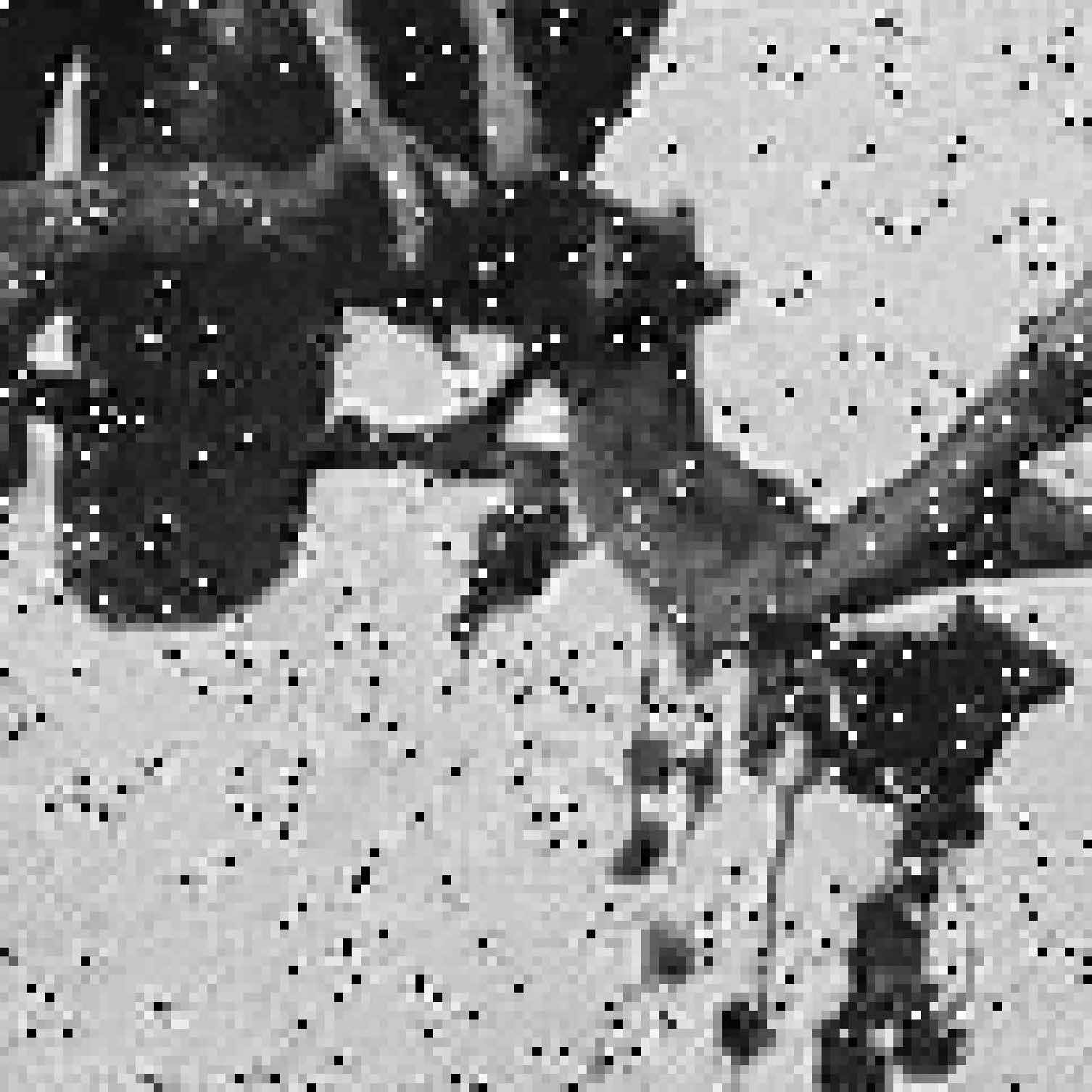} &
		\includegraphics[width=\imgsizezoom\linewidth]{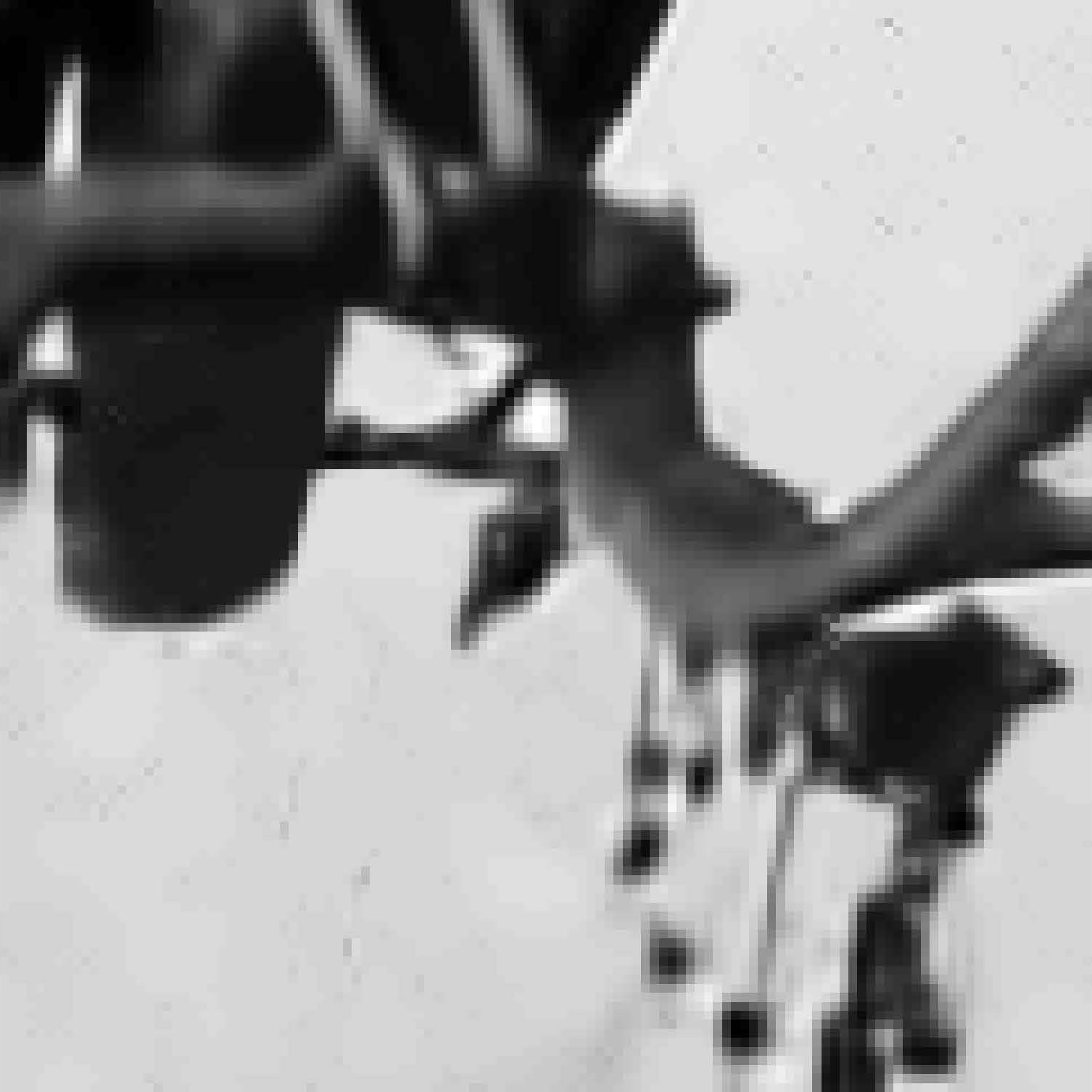} \\
     &  0.8989 & 0.8832 & 0.2822 & 0.8963  \\[0.2cm]
     CS &  MF & NLM patch  & \textbf{\textit{LCD}} & \textbf{\textit{NLCD}} \\
    \includegraphics[width=\imgsizezoom\linewidth]{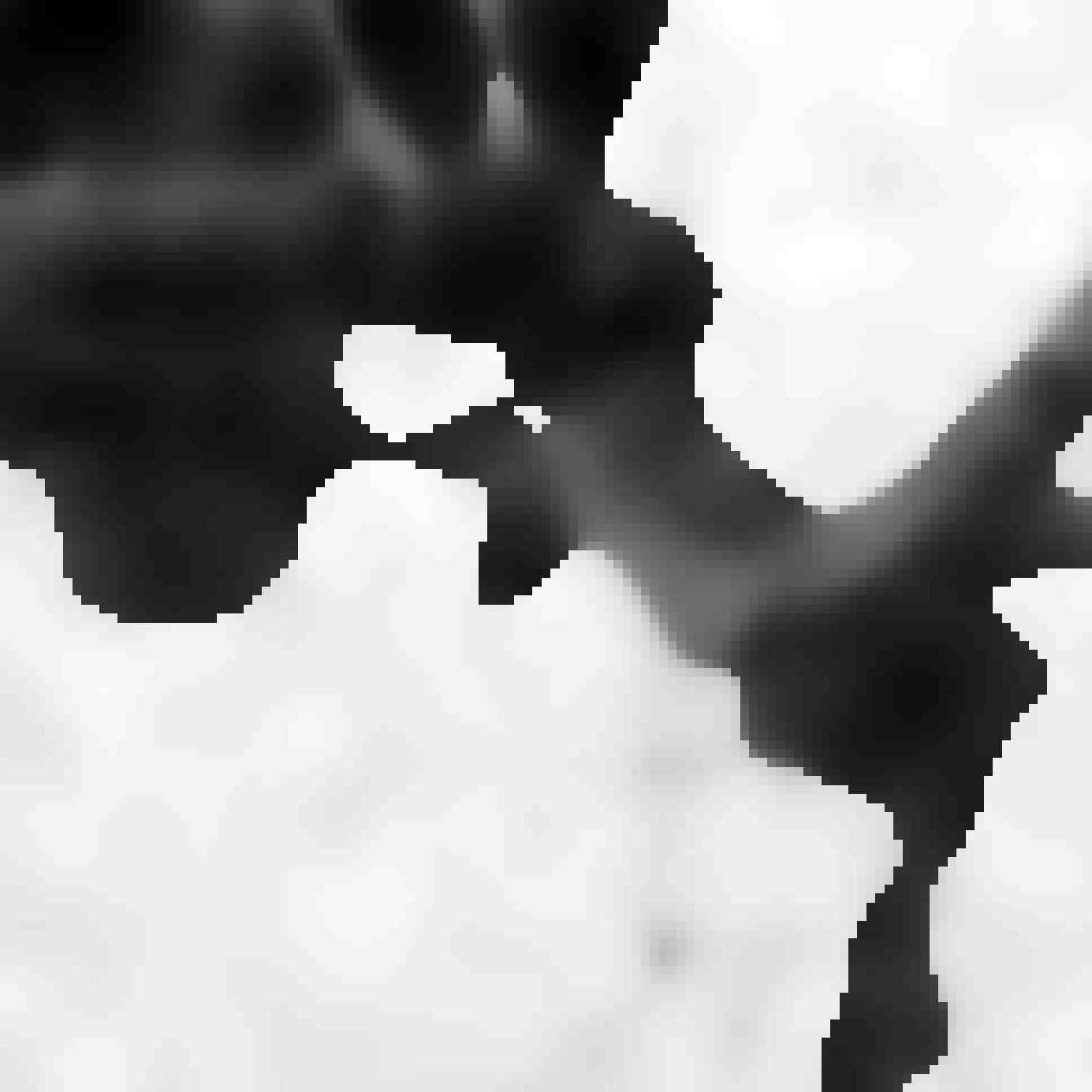}  &
    \includegraphics[width=\imgsizezoom\linewidth]{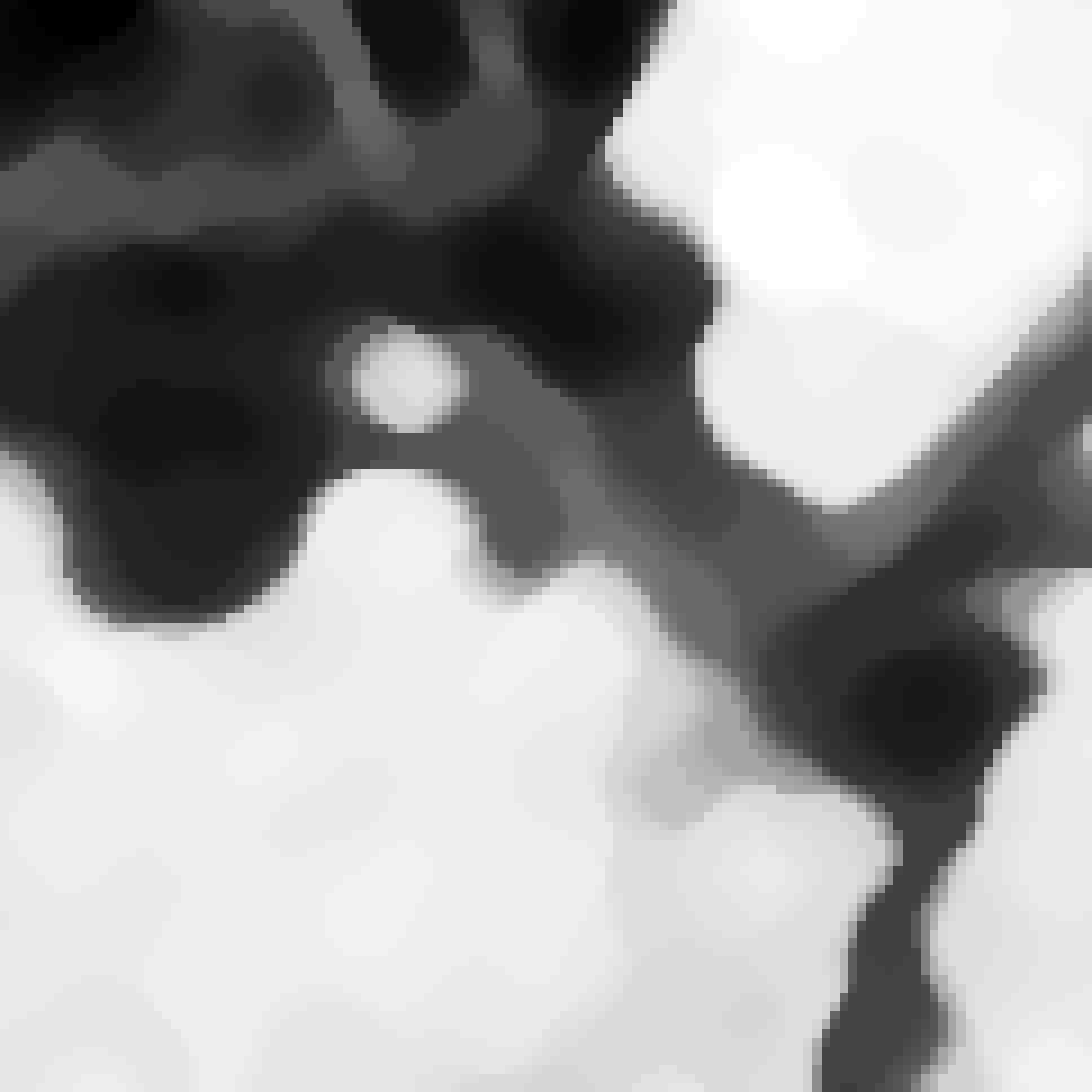} &
		\includegraphics[width=\imgsizezoom\linewidth]{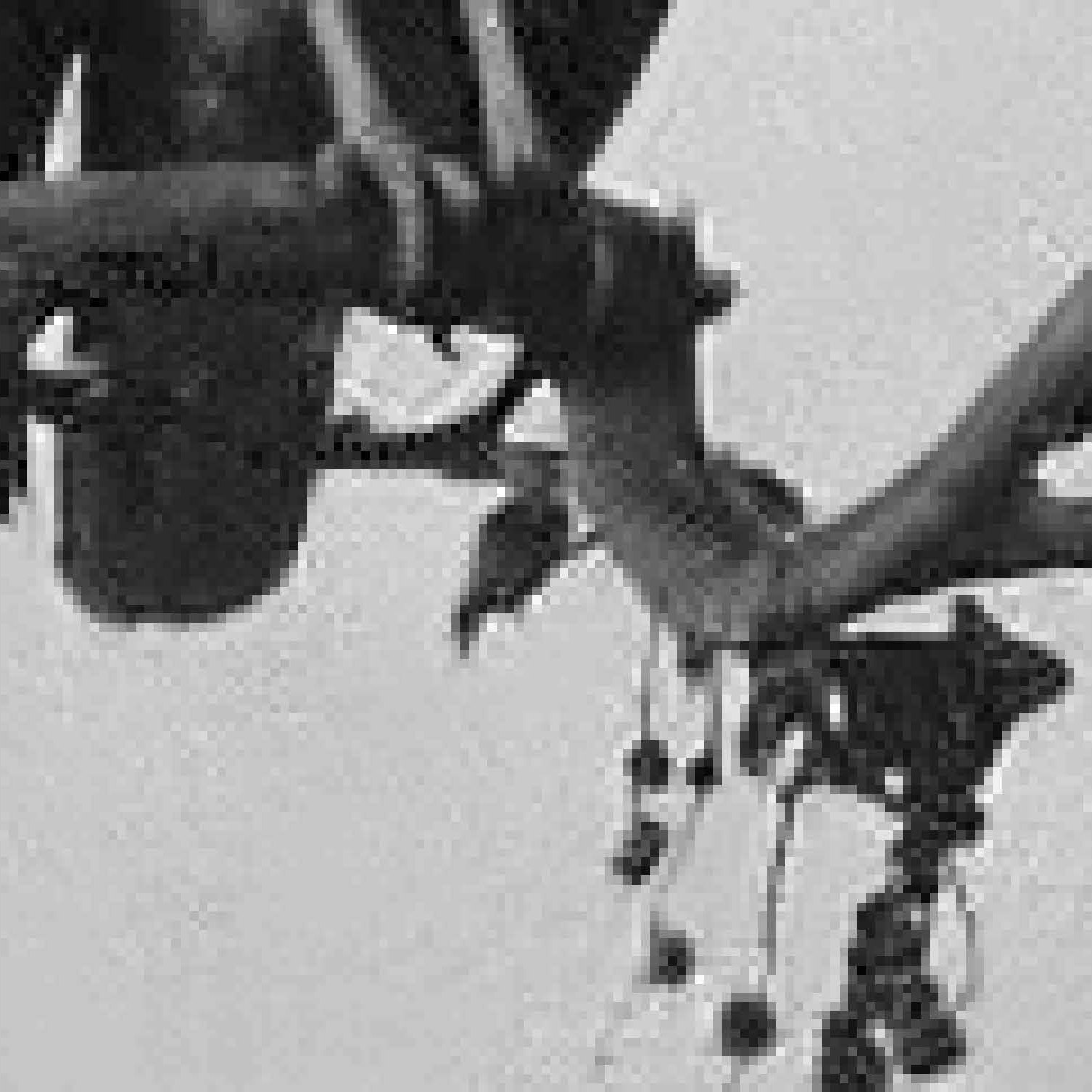}  &
    \includegraphics[width=\imgsizezoom\linewidth]{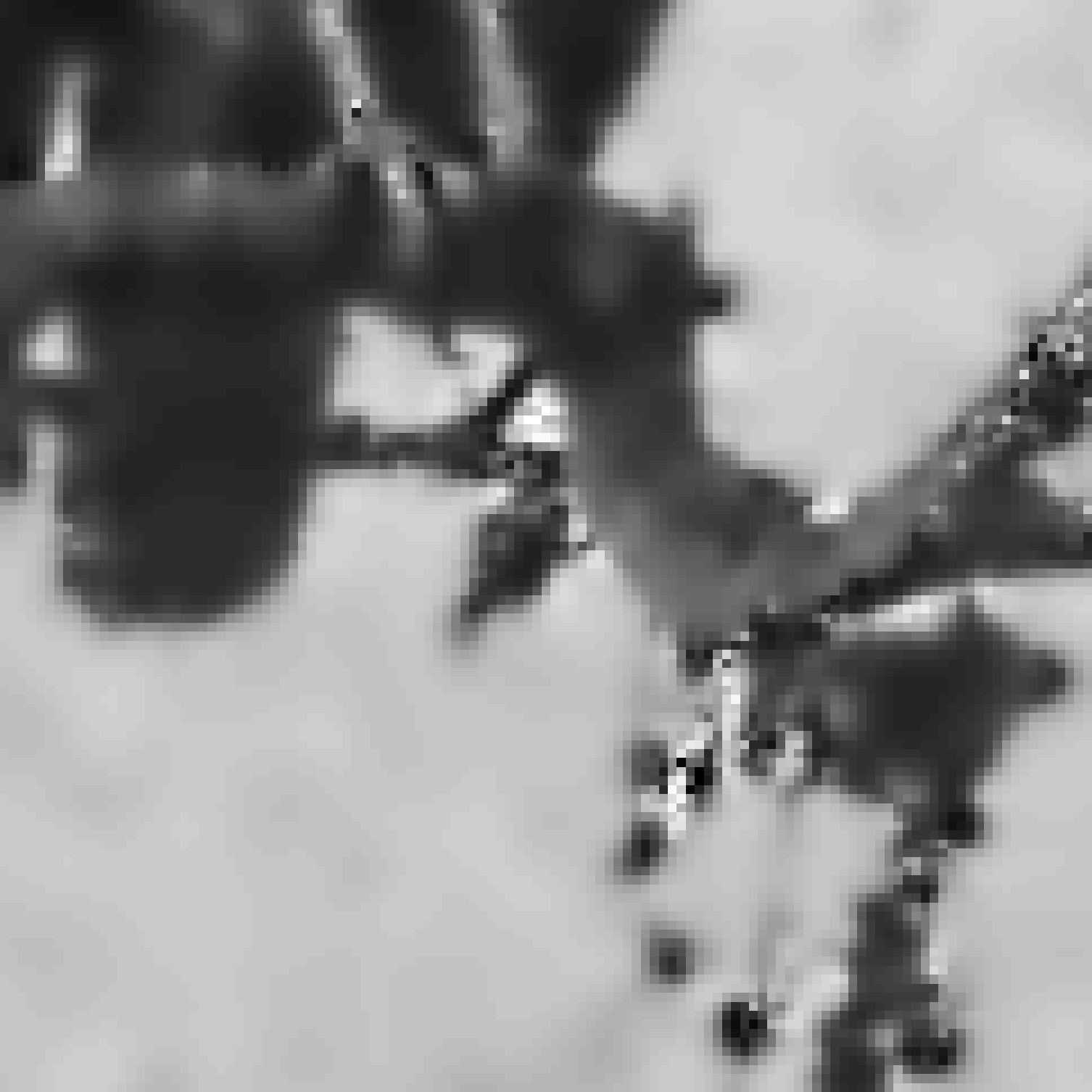}  &
    \includegraphics[width=\imgsizezoom\linewidth]{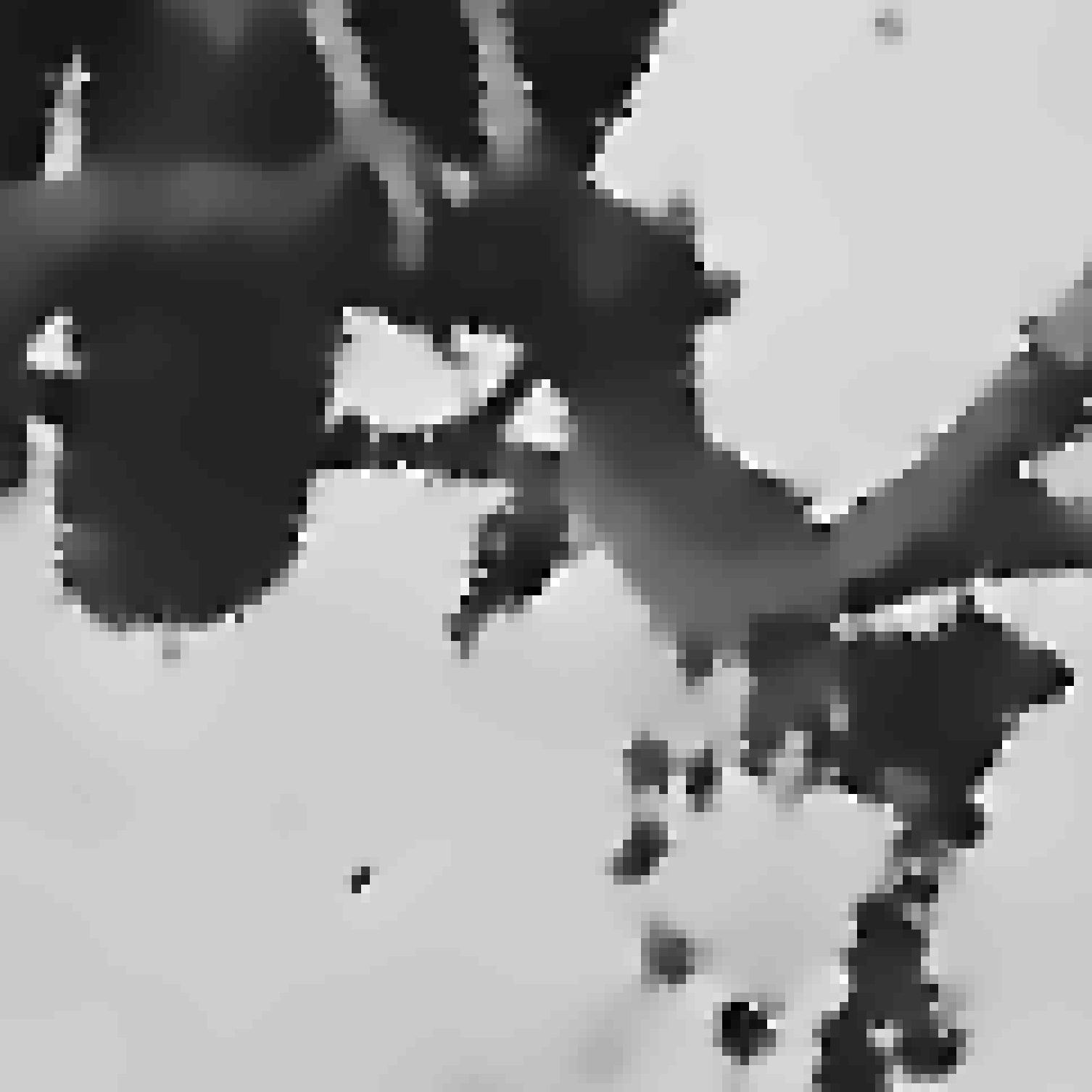} \\
    0.8694 &  0.8632 & 0.8329 & 0.8705 & 0.9028 \\[0.2cm]   \end{tabular} \\[-0.3cm]
   %%%%%%%%%%%%%%%%%%%%%%%%%%%%%%%%%%%%%%%%%%%%%%%%%%%%%%%%%%%%%%%%%%%%%%%%%%%%%%%%%%%%%%
   \begin{tabular}{c}     
              Boat\&House            \\[0.12cm]                        
     \multicolumn{1}{c}{\includegraphics[width=\imgsize\linewidth]{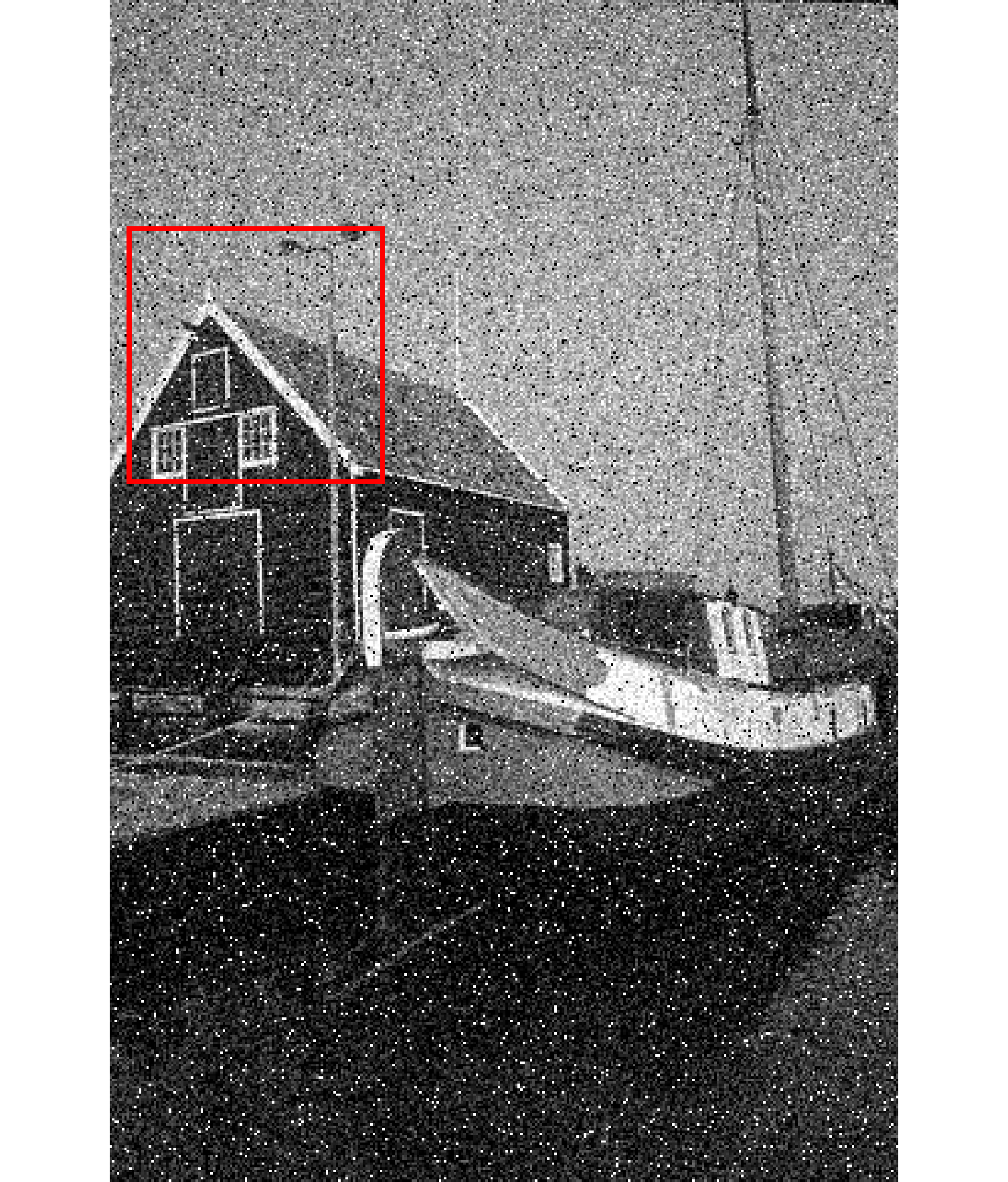}}  \\
     Initial noise = 0.1687 \\[0.2cm]
   \end{tabular}
   % ---------------------
   \begin{tabular}{ccc ccc}                             
      Original &  NLD & AD & NLM pixel & BM3D  \\
    \includegraphics[width=\imgsizezoom\linewidth]{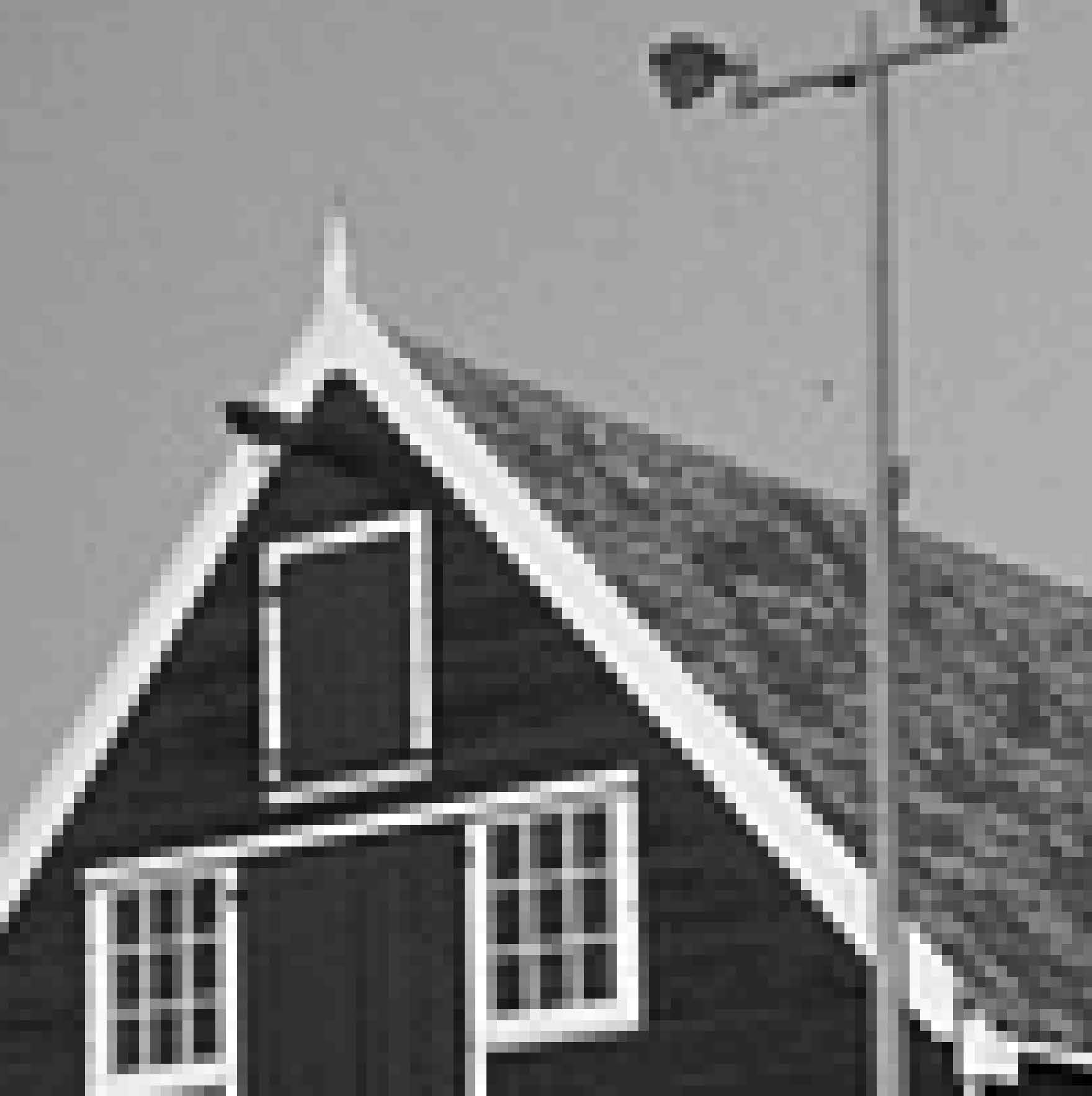}  &
    \includegraphics[width=\imgsizezoom\linewidth]{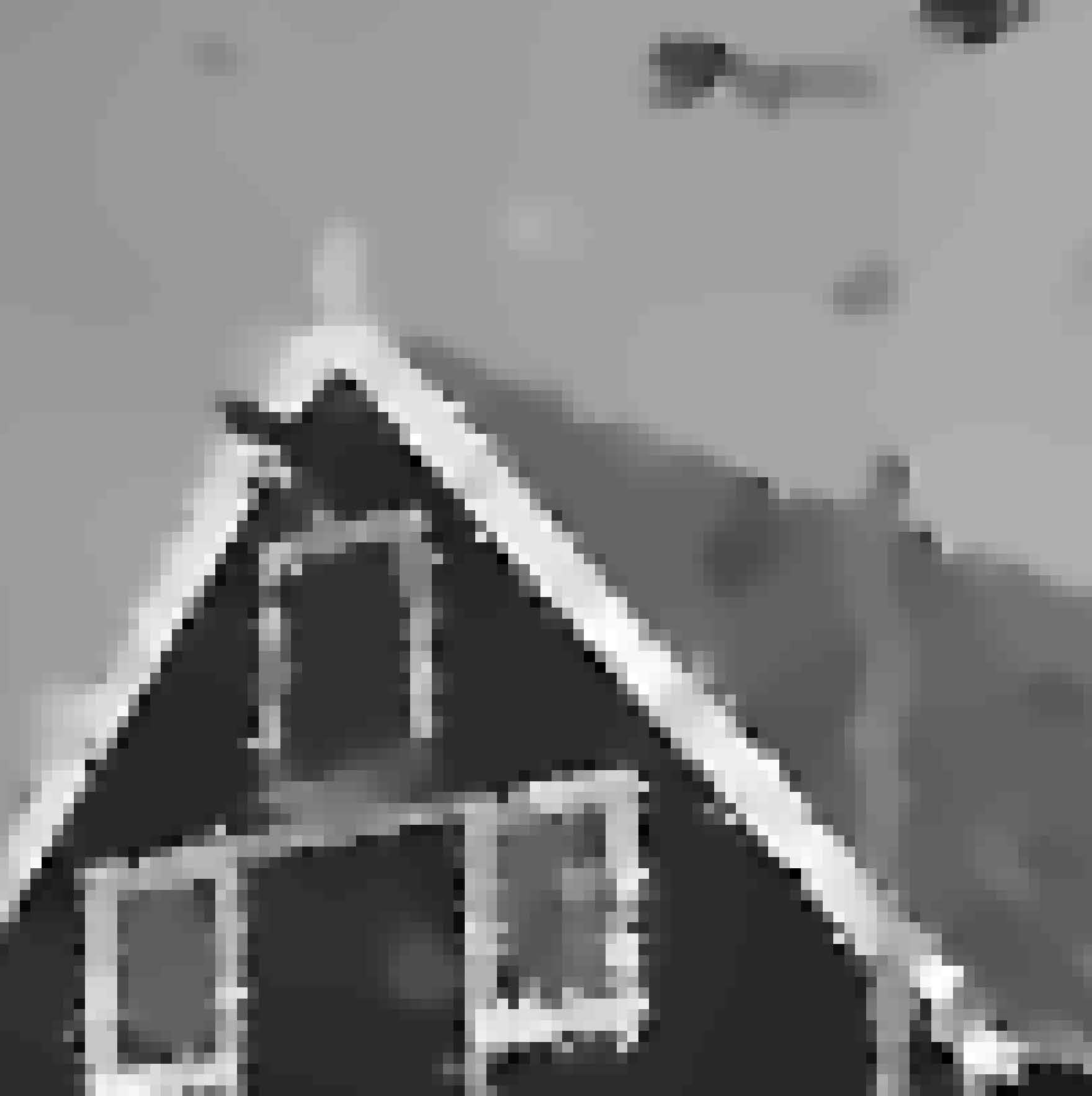}  &
    \includegraphics[width=\imgsizezoom\linewidth]{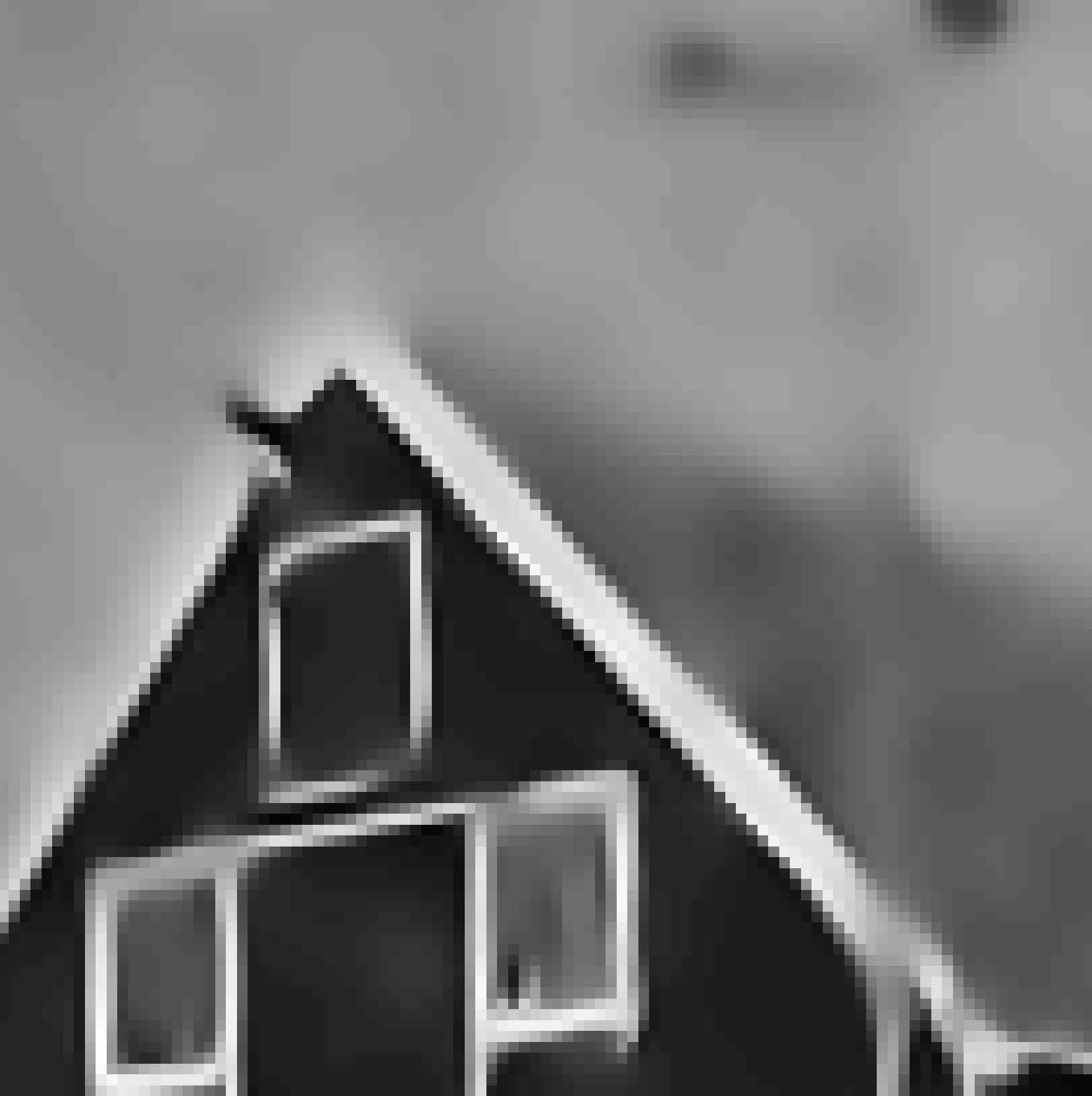}  &
    \includegraphics[width=\imgsizezoom\linewidth]{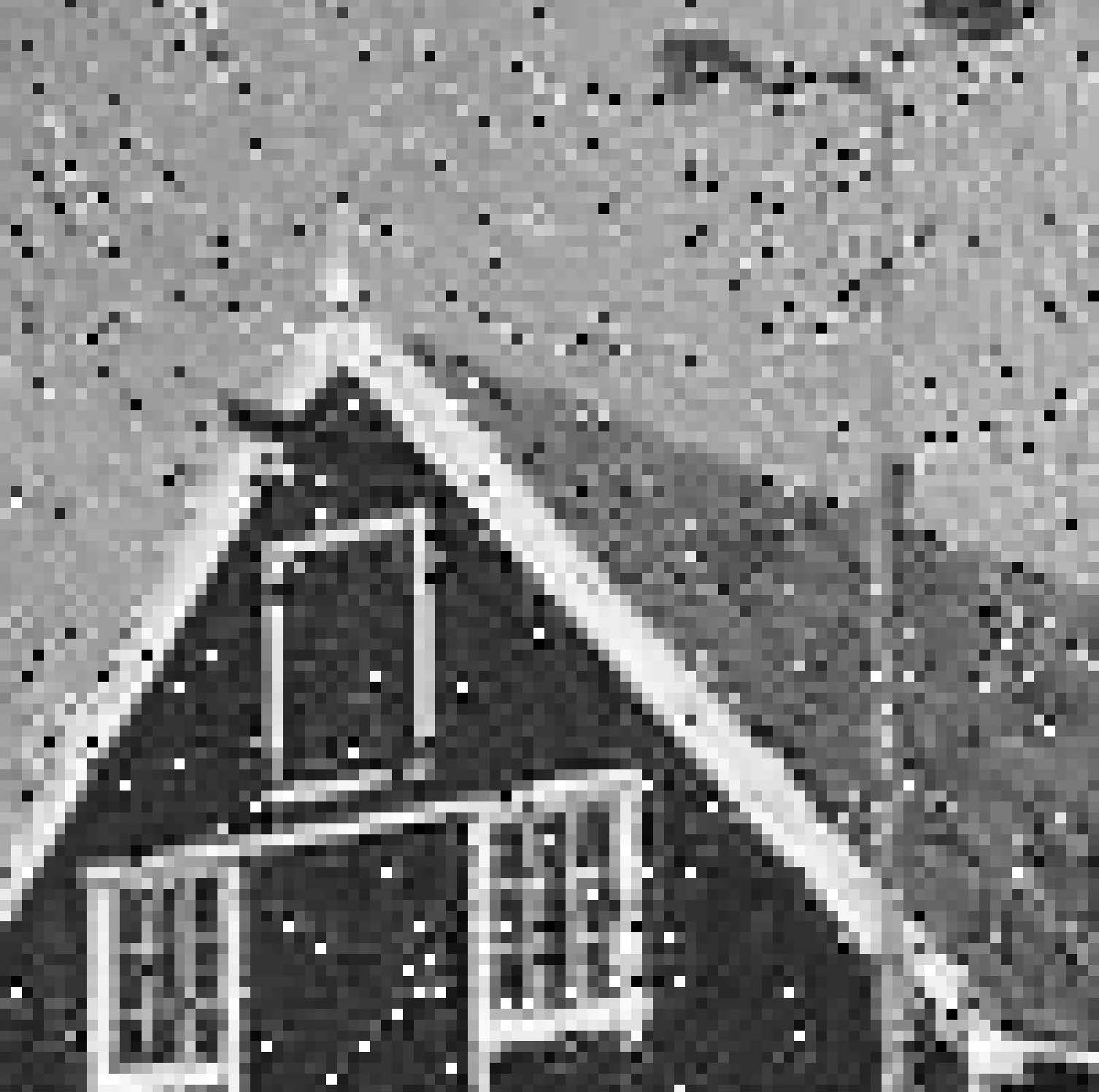} &
		\includegraphics[width=\imgsizezoom\linewidth]{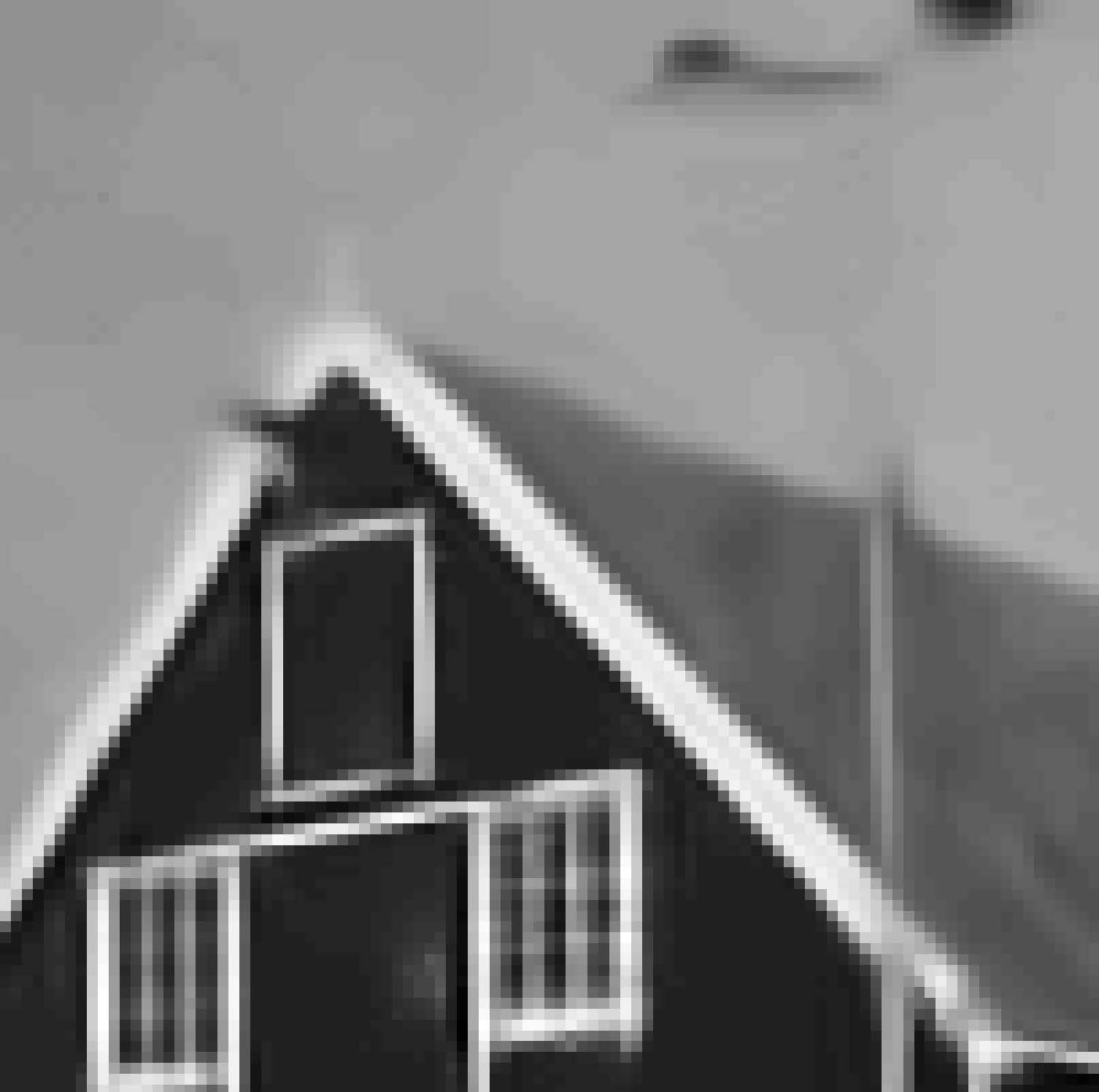} \\
     &  0.7511 & 0.7380 & 0.2964 & 0.7624  \\[0.2cm]
     CS &  MF & NLM patch  & \textbf{\textit{LCD}} & \textbf{\textit{NLCD}} \\
    \includegraphics[width=\imgsizezoom\linewidth]{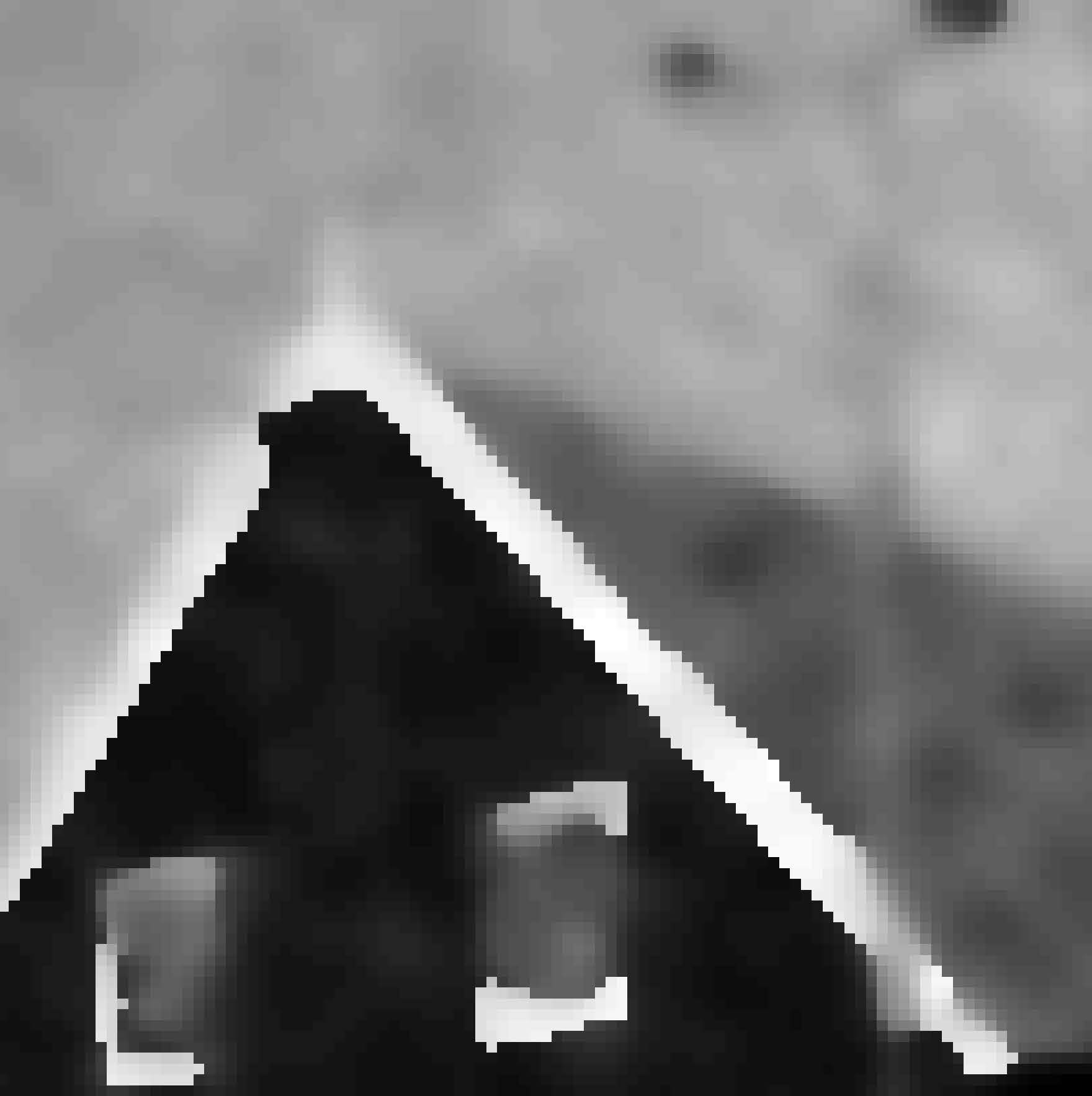}  &
    \includegraphics[width=\imgsizezoom\linewidth]{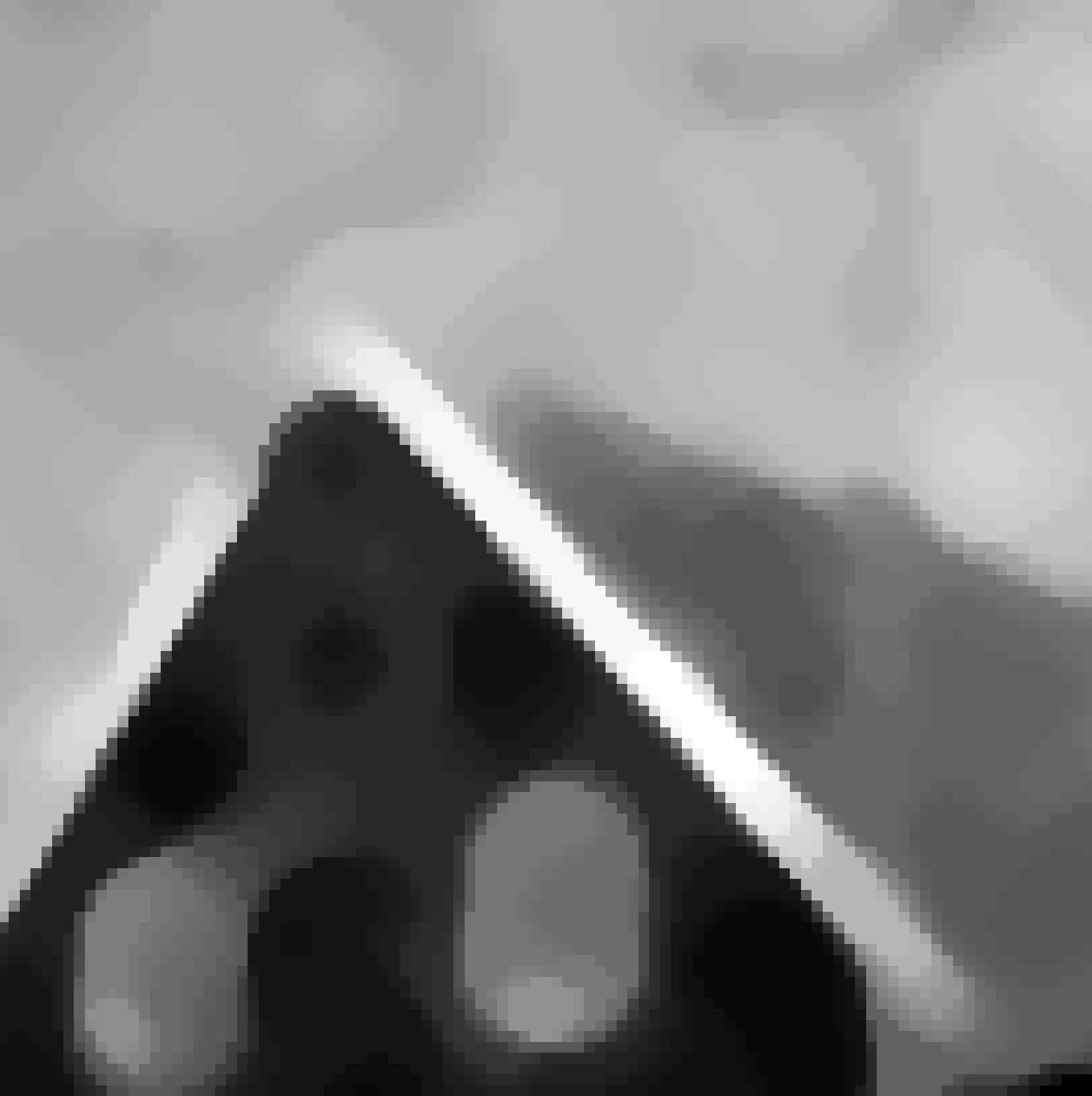} &
		\includegraphics[width=\imgsizezoom\linewidth]{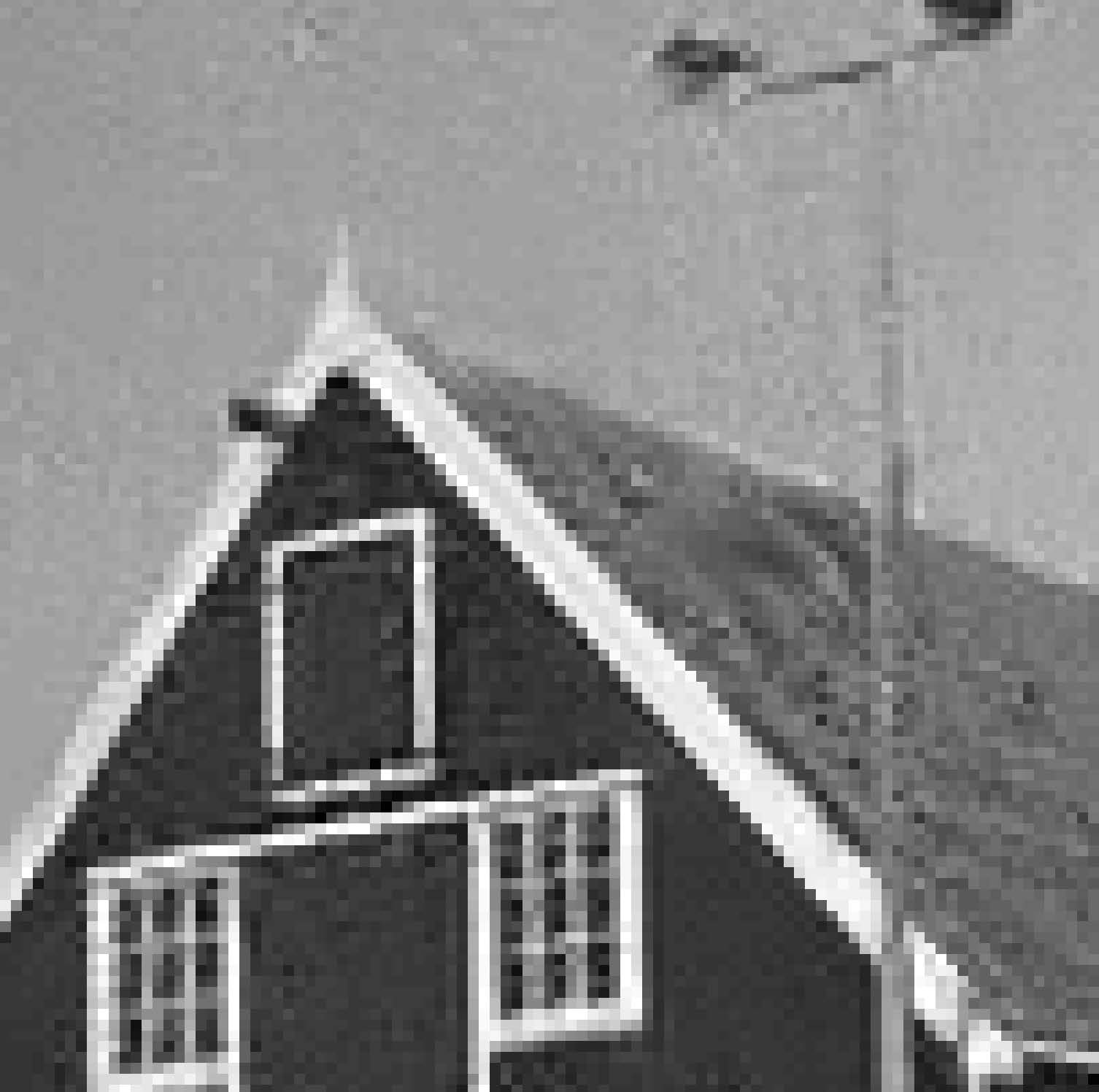}  &
    \includegraphics[width=\imgsizezoom\linewidth]{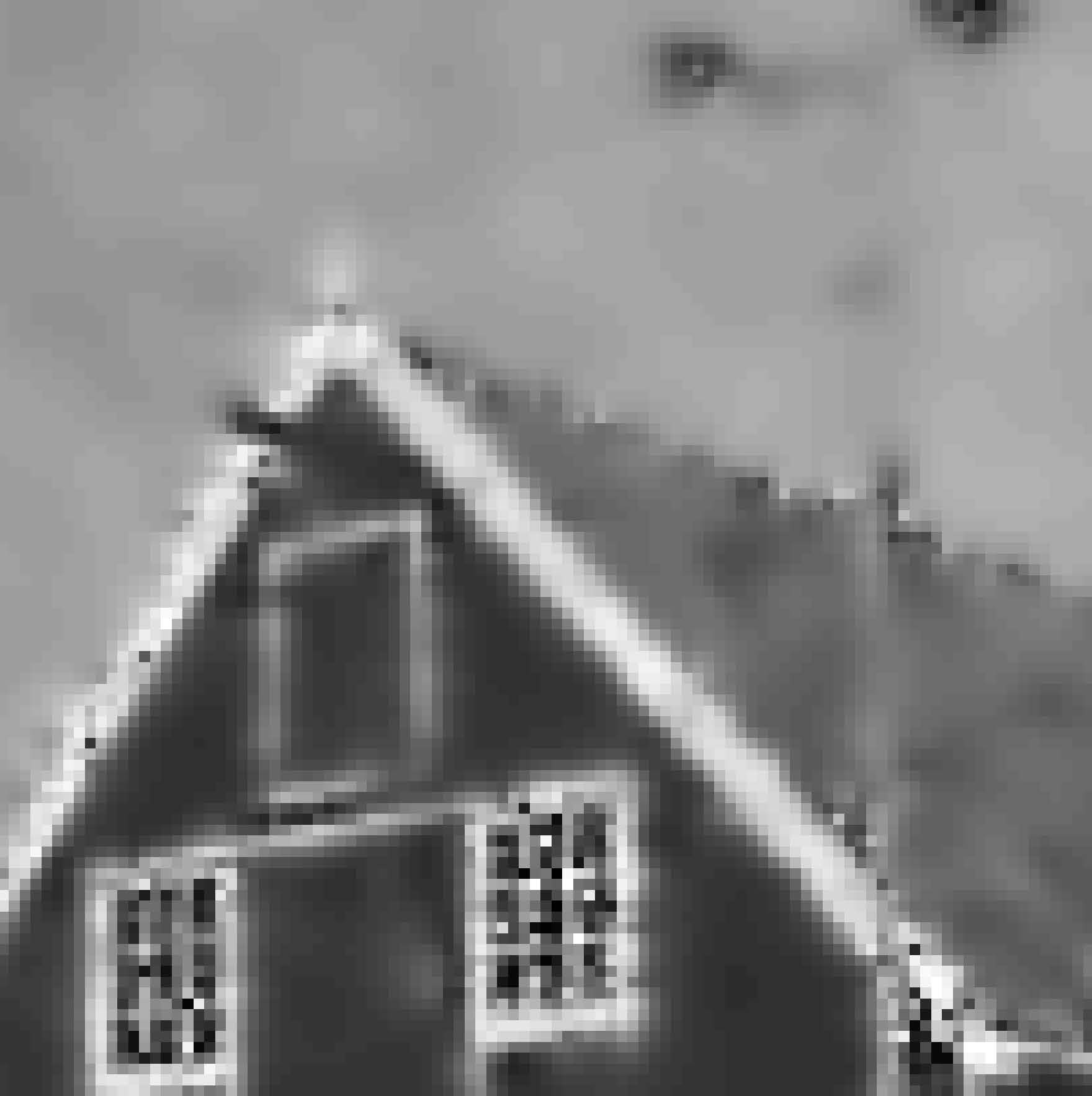}  &
    \includegraphics[width=\imgsizezoom\linewidth]{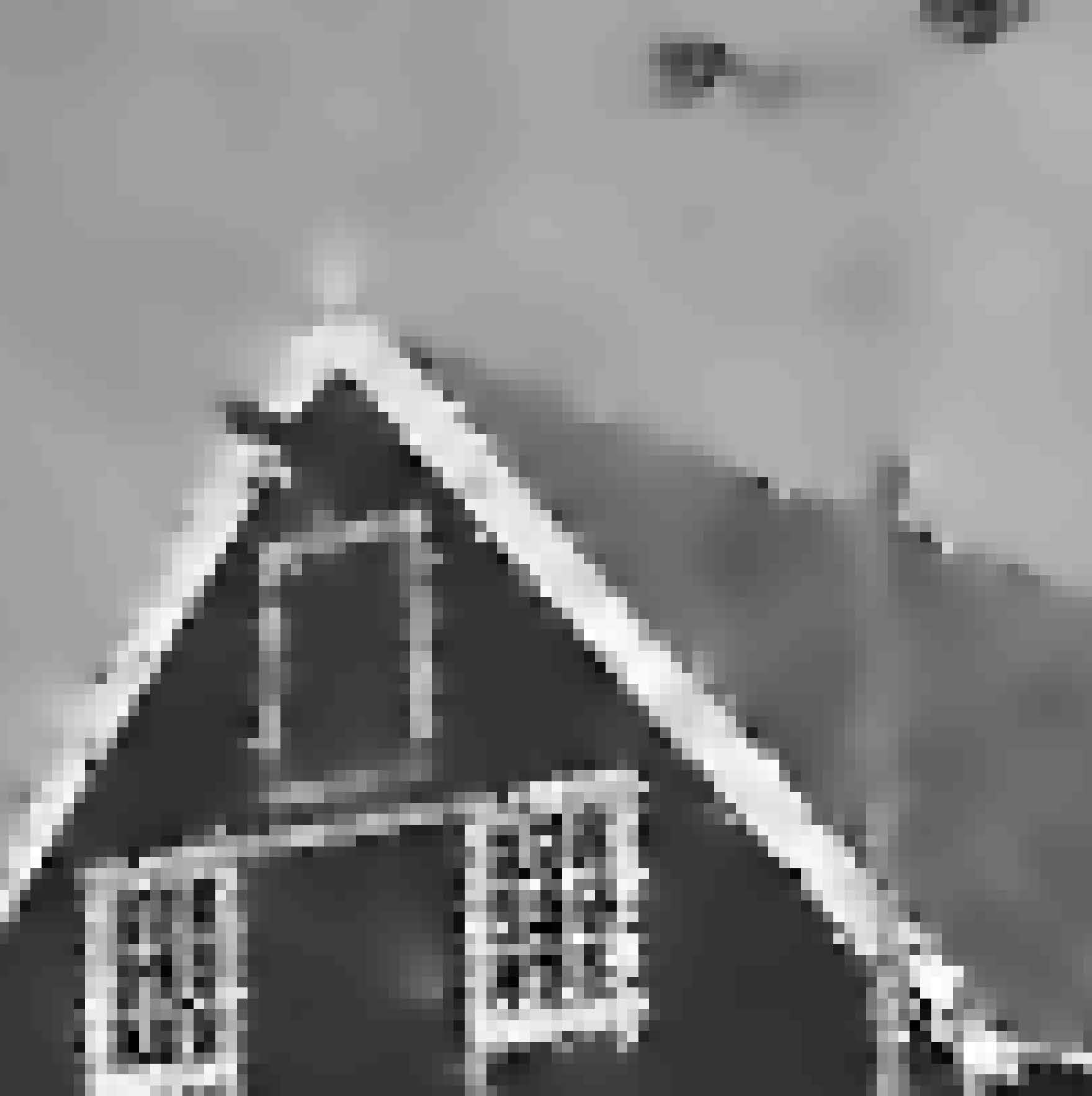} \\
    0.7269 &  0.7242  & 0.7798 & 0.7481 & 0.7665 
   \end{tabular} \\[-0.3cm]
\end{tabular}
\end{centering}   
\caption{Visualization of certain parts of different images (Cameraman, Hawk, Boat\&House) of noise level $\sigma = 30$. A visual comparison is done between the introduced methods. 
} 
\label{fig:denosingvisu}     
\end{figure*}

\subsection{Results}
In Figure \ref{fig:denoising} we show the best obtained SSIM value for each image and considered noise level. The pixel based NLM variant \cite{NLMpixel} has SSIM values between 0.3 and 0.5 for all images as it poorly reduces impulse noise. To focus on higher SSIM values we do not show values for pixel NLM.

Generally, best denoising results are obtained by the patch version of the non-local means, especially for low gaussian noise levels. BM3D also gains high SSIM. For low Gaussian noise CS shows good results as well. This was expected as channel smoothing, as well as median filtering, are well suited methods for removing clear outliers. The NLCD gives quite high SSIM for higher Gaussian noise levels.

For few special cases we observe that NLCD performs best, even better than the state-of-the-art denoising methods BM3D and NLM, which is unexpected, see e.g. result for the hawk image. In all cases NLCD is comparable and most times better than the other diffusion-based methods, especially well \eg in the skyline image or the cameraman. As soon as Gaussian noise of around $\sigma = 15$ or $\sigma = 20$ has been added to the image the NLCD also ouperforms CS. 

The aim of this work was not to construct the best denoising algorithm but to combine the advantages of channel smoothing and diffusion based schemes. The NLCD outperforms in all cases CS if a medium or high amount of Gaussian noise is present and it outperforms in most cases NLD and AD for all noise levels. In some cases NLCD even shows competitive results compared to state-of-the-art denoising methods. 

In Figure \ref{fig:denosingvisu} a visualization of certain image close ups can be seen. The CS and MF show a comic like behavior, maintaining the main edges well but details are lost as can be seen in the face or the camera. The pixel based NLM cannot handle the impulse noise at all. Better preservation of details is achieved using AD or NLD. Good results show the patch based NLM and BM3D. However, for patch NLM the image still shows some Guassian noise and for BM3D still some details are lost. The proposed LCD show an improvement compared to CS and MF, but due to its linear form the edges are smeered. Visually, better results compared to CS and diffusion methods are obtained with NLCD. It keeps the details and it is able to handle impulse noise as well as Gaussian noise.

\vspace{-4mm}
\section{Conclusions}
For the aim of filtering noisy images a new linear and a new non-linear diffusion scheme has been presented using advantages of channel representations. For this purpose we derived an iterative filtering scheme by minimizing a corresponding energy functional. Including the channel framework leads to a robust filtering well suited for images corrupted with Gaussian as well as impulse noise. We analysed the denoising behaviour of the proposed method on commonly used scalar valued images and compared the methods to similar as well as state of the art methods. It turned out that the new method outperforms the other diffusion-based methods if impulse noise and a medium or high amount of Gaussian noise are present. In some cases it even outperforms state-of-the-art denoising methods. In future investigations, application of the novel NLCD scheme to DTMRI and HARDI data may be of interest.

\vspace{-4mm}
\section*{Acknowledgements}

\noindent This research has been in part supported by the Swedish Research Council through a grant for the project \emph{Visualization-adaptive Iterative Denoising of Images} 
and has received in part funding from the European Communitys Seventh Framework Programme FP7/2007-2013
Challenge 2 Cognitive Systems, Interaction, Robotics under grant agreement No
247947 GARNICS.

\bibliographystyle{apalike}

\vspace{-4mm}
{\small
\bibliography{example}}

\end{document}